# Evaluating the Determinants of Mode Choice Using Statistical and Machine Learning Techniques in the Indian Megacity of Bengaluru

Tanmay Ghosh[1] and Nithin Nagaraj[2]

[1]Energy, Environment, and Climate Change Program, National Institute of Advanced Studies (NIAS), Indian Institute of Science Campus, Bengaluru, Karnataka 560012, India

[2]Consciousness Studies Programme, National Institute of Advanced Studies (NIAS), Indian Institute of Science Campus, Bengaluru, Karnataka 560012, India

Email(s): tanmay@nias.res.in, nithin@nias.res.in

## Abstract

The decision making involved behind the mode choice is critical for transportation planning. While statistical learning techniques like discrete choice models have been used traditionally, machine learning (ML) models have gained traction recently among the transportation planners due to their higher predictive performance. However, the black box nature of ML models pose significant interpretability challenges, limiting their practical application in decision and policy making.

This study utilised a dataset of 1350 households belonging to low and low-middle income bracket in the city of Bengaluru to investigate mode choice decision making behaviour using Multinomial logit model and ML classifiers like decision trees, random forests, extreme gradient boosting and support vector machines. In terms of accuracy, random forests model performed the best (0.788 on training data and 0.605 on testing data) compared to all the other models.

This research has adopted modern interpretability techniques like feature importance and individual conditional expectation plots to explain the decision making behaviour using ML models. Key findings indicate that travel cost and travel time contribute highest to the model's predictive output, across all ML models. A higher travel costs significantly reduce the predicted probability of bus usage compared to other modes (a 0.66% and 0.34% reduction using Random Forests and XGBoost model for 10% increase in travel cost). However, reducing travel time by 10% increases the preference for the metro (0.16% in Random Forests and 0.42% in XGBoost).

This research augments the ongoing research on mode choice analysis using machine learning techniques, which would help in improving the understanding of the performance of these models with real-world data in terms of both accuracy and interpretability.

## 1. Introduction

Urban transportation infrastructure is the backbone of modern cities economy. With the growing need for travel, the transportation infrastructure is undergoing rapid changes. Given the capital intensive nature of these infrastructure projects, it is necessary to plan these projects by understanding the needs and travel behaviour of the commuters with a high degree of accuracy. Planning a new infrastructure project entails analysis of all the causal factors influencing commuters' mode choice decision. Among the causal factors are the basic characteristics of the transportation system like travel cost and travel time to reach from the origin to destination, socio-economic characteristics of the commuters like income, number of vehicles owned by the household etc. as well the built environment characteristics of the residential and destination location. As the conventional approaches which has focussed on adding the capacity of the transportation network is no longer feasible and sustainable, it is important to manage travel demand by encouraging commuters to adapt their travel behaviour (Batur and Koc, 2017). This necessitates a deeper understanding of the random variations in the travel behaviour of the commuters which not only minimises forecasting error pertaining to the decisions involved in choosing a mode of travel but also informs the causality involved in mode choice decisions. The transportation policy makers endeavour to understand the travel behaviour of the potential commuters while planning for investments in transportation projects, the implication of which is an additional cost to the society and the economy.

Current approaches to simulate and forecast mode choice makes use of the econometric modeling techniques like Random Utility Models (RUM) (McFadden, 1979) which are based on the economic theory of utility maximization, which assumes that the commuter is a perfect rational entity whose decision to choose a modal alternative depends on the utility derived from using that particular alternative. As such, the commuter will choose the alternative that yields the highest utility among the set of available alternative choices. The discrete choice models are special collection of models which follows the random utility principles and have a solid economic and statistical foundations and hence, require the dataset to fulfil a specific set of assumptions. The common critique of these models is the statistical assumption of the property of independence of irrelevant alternatives (IIAs), which implies that the effects of alternative attributes are compensatory and result in biased estimates and incorrect predictions in cases that violate the IIA property (Koppleman and Wen, 1998), although significant improvements for elimination of the IIA property have been made. Their predetermined structures may often misestimate or ignore partial relationships between explanatory variables and alternative choices for specific subgroups in a population. The additive nature of utility functions may not adequately model the correlation property between predictor variables and the dependent variables (Cheng et al., 2014). Along with the statistical critique, the assumption of perfect rationality of the decision makers is problematic (Miljkovic, 2005). The randomness in real world decision due to various endogenous factors like cognitive biases, where individuals favour information that aligns with their pre-existing beliefs and exogenous factors, such as social influences, where societal norms, cultural values, and peer pressures can drive individuals to make choices that align more with external expectations than their own intrinsic preferences, or information asymmetry, where everyone do not have access to the same quality and quantity of information, eludes the narrow confines of rationality.

Unlike traditional choice models, Machine Learning (ML) models do not rely on inherent economic theory to make a prediction. These models are inherently data-centric and attempt to learn the underlying patterns and data distributions for making predictions agnostically (Cranenburg et al., 2022). The adoption of new transportation technologies like use of contactless smart cards, vehicle tracking cameras, use of Global Positioning System (GPS) to live track the location of the vehicles, have increased the data harnessing capabilities (Bieler et al., 2022). This provides the opportunity to build ML models with greater capability to model passenger travel behaviour and infer the causal relationship between commuter's decision making and attributes of the transportation system, socio-economic characteristics and environmental conditions. (Hillel et al., 2021)

Recently, the transportation community has started shown increasing interest in harnessing the predictive accuracy of machine learning algorithms to address various transport-related research problems, which includes forecasting mode choice behavior, traffic forecasting, and traffic safety analysis. Contemporary studies have highlighted the notable improvements achieved through the application of machine learning techniques in modeling travel mode choice behavior, outperforming traditional discrete choice models (Xie and Parkany, 2003; Wang and Ross, 2018; Cranenburg et al., 2022).

The objective of the current study is to investigate the mode choice decision of the commuters by analysing travel behaviour data collected through a field survey. This study compares the relative predictive performance of the machine learning classification algorithms, as well as traditional discrete choice model like Multinomial Logit (MLN). While MNL models traditionally rely on their inferential capability to interpret the causal factors involved in decision making, this study tries to employ the current developments in the interpretable machine learning landscape. By using tools like feature importance and conditional expectation plots, we aim to combine the model's predictive ability with its interpretability, thereby improving our understanding of complex decision-making processes. This study is unique due to its utilisation of field based travel survey data, which is inherently noisier than synthetic datasets, to compare the performance of classifiers. To the best of the authors' knowledge, studies based on field survey data are rarely represented in contemporary literature on machine learning research. Consequently, the findings of this research would add value for both the transportation modelers and machine learning community and would offer insights on how to employ appropriate classifier suites in the endeavour to model high dimensional datasets, with seemingly non-linear relationships between explanatory features and the mode choice.

This paper is organized as follows. Section 2 reviews the previous literatures which have employed machine learning algorithms to solve mode choice problem. Section 3 describes the study area and the travel behaviour data used for modeling mode choice. Section 4 explains the methodological framework of the modeling technique (both MNL and ML). Section 5 outlines the experimental framework used to specify the models. Section 6 details the findings and discusses the results and finally, Section 7 concludes the study, highlighting its relevance in policymaking, addressing the limitations of the study as well as suggesting scope for the future research.

## 2. Background

Empirical investigations consistently demonstrate the superior performance of ensemble methods and Support Vector Machines (SVMs) compared to the Multinomial Logit (MNL) model in mode choice modeling (Zhao et al., 2020; Garcia et al., 2022; Omrani, 2015; Hagenauer and Helbich, 2017). These studies serve as benchmarks for future approaches in this field. Among the ensemble methods, Gradient Boosting Decision Trees (GBDT) has gained prominence, with the widely recognized XGBoost algorithm, implementing GBDT, receiving significant attention due to its superior predictive performance (Wang and Ross, 2018). Extensive evidence supports the overall superiority of XGBoost over the MNL model in terms of predictive accuracy. However, it is important to note that XGBoost presents challenges in terms of interpretability.

To elucidate mode choice determination, the Random Forest algorithm has been employed. Ermagun et al. (2014) conducted an analysis of travel mode choice using Random Forest and Nested Logit models, incorporating socio-economic attributes, built environment variables, and psychological variables as explanatory factors. The results revealed the significant superiority of the Random Forest model in terms of prediction accuracy compared to the nested logit model. Similarly, Cheng et al. (2019) utilized socio-demographic attributes, trip information, built environment variables, and various machine learning models such as Random Forest, AdaBoost, and SVM, along with econometric models like the Multinomial Logit model (MNL), to analyse and predict mode choice. Among all the models, Random Forest exhibited the highest prediction accuracy and computational efficiency (Hagenauer & Helbich, 2017)

In the realm of extracting behavioral insights from these models, the scholarly works of Ramsey and Bergtold (2021) and Wang et al. (2020, 2021b) hold significant importance. These studies provide valuable methodologies for uncovering and analysing behavioral information within the examined models, offering researchers a comprehensive understanding of the unique advantages and behavioral implications of Deep Neural Networks (DNNs).

Neural networks are susceptible to overfitting as these models have the capacity to learn complex patterns which also includes the noise in the training data and, hence are not able to generalise very well to the external dataset, leading to the problem of overfitting Additionally, a large amount of data is required to train a neural network to avoid overfitting and to ensure generalisability (Bejani and Ghatee, 2021; Li et al., 2019).

## 3.Data

### 3.1 Study area

This study investigates the travel behavior of households residing in the vicinity of metro stations (1500 metre radius) in the Bruhat Bengaluru Mahanagara Palike (BBMP) region of Bengaluru, which also serves as the city's central zone. Encompassing an area of 710 square kilometres, the BBMP region is inhabited by a population of 8.44 million, including a working population of 3.69 million based on the 2011 Census. Projections indicate a projected population increase to 11.34 million by 2018 and 14.3 million by 2025. According to a feasibility study by Rail India Technical and Economic Service (RITES) in 2019, approximately 32% of daily trips for various purposes within the region are undertaken using public transport, which encompasses the Bangalore Metropolitan Transport Corporation (BMTC) bus service and the Bangalore Metro Rail Corporation Limited (BMRCL) metro service. Non-motorized transport modes, such as walking and cycling, account for 26% of daily trips, while the remaining population relies on motorized modes of transportation, with 7% utilizing cars, 27% utilizing two-wheelers, and the rest opting for intermediate public transport options such as autos and taxis (RITES, 2019). The spatial extent of the study area is depicted in Figure 1.

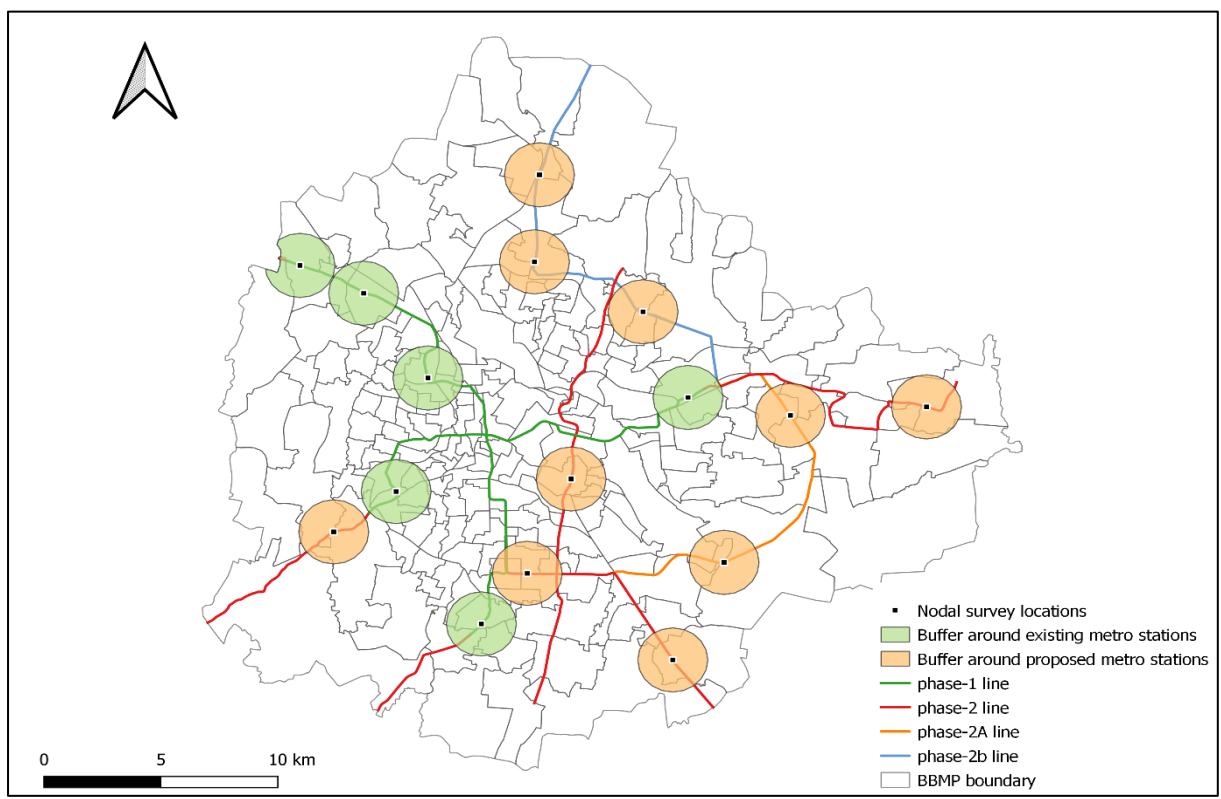

**Figure 1: A schematic description of the study area.**

This study aims to analyse the travel patterns and preferences of households located near metro stations, considering both the operational Phase 1 and the upcoming Phase 2 (including 2A and 2B) of the Bengaluru metro. Travel behaviour is impacted by various factors like the attributes of the transportation modes available to the population, socio-economic and demographic characteristics of the population, as well as the built environment features of the urban area where the population is residing and traveling for work, recreation, schooling etc.

Table 1 presents the variables used in the dataset and Table 2 details the class labels against the number of instances.

**Table 1. Variables used in the final dataset. These ten variables constitute the raw original features for the prediction task.**

| Variable | Description | Type | Mean/mode |
|---|---|---|---|
| Household attributes | | | |
| Household vehicle ownership | Household ownership two wheeler | Continuous | |
| Age | Age of the respondents | Continuous | 32 |
| Total Household Income (Indian rupees) | Household income | Continuous | 7730 |
| Gender of the respondent | Gender of the respondents | Categorical (Male =0, Female =1) | 1 |
| Travel attributes | | | |
| Travel distance (metres) | Total Distance between origin and destination | Continuous | 5462.5 |
| Trip time (minutes) | Time taken to complete one trip between origin and destination | Continuous | 29 |
| Trip cost (Indian rupees) | Total cost incurred to complete one trip between origin and destination | Continuous | 9 |
| Built Environment | | | |

| Availability of metro | Availability of metro at the origin end of the journey | Categorical | 0 |
|---|---|---|---|
| Population density(1/sq.km.) | Population density of the ward | Continuous | 19063 |
| Employment density (1/sq.km.) | Employment density of the destination ward | Continuous | 8151 |
| mode class[*] | | Categorical | 8 |
| [*] Target variable | 1- metro, 2- bus, 3 -SR, 4-Auto, 5-Two-wheeler, 6-Car, 7-cycle, 8-walk | | |

**Table 2. Class labels and number of instances.**

| Class label | Class name | Number of instances |
|---|---|---|
| 1 | Metro | 79 |
| 2 | Bus | 1018 |
| 3 | Shared ride | 317 |
| 4 | Auto | 238 |
| 5 | Two wheeler | 988 |
| 6 | Four wheeler | 98 |
| 7 | Cycle | 31 |
| 8 | Walk | 1625 |

## 4. Methods

This section presents the existing and new emerging approaches used for mode choice predictions.

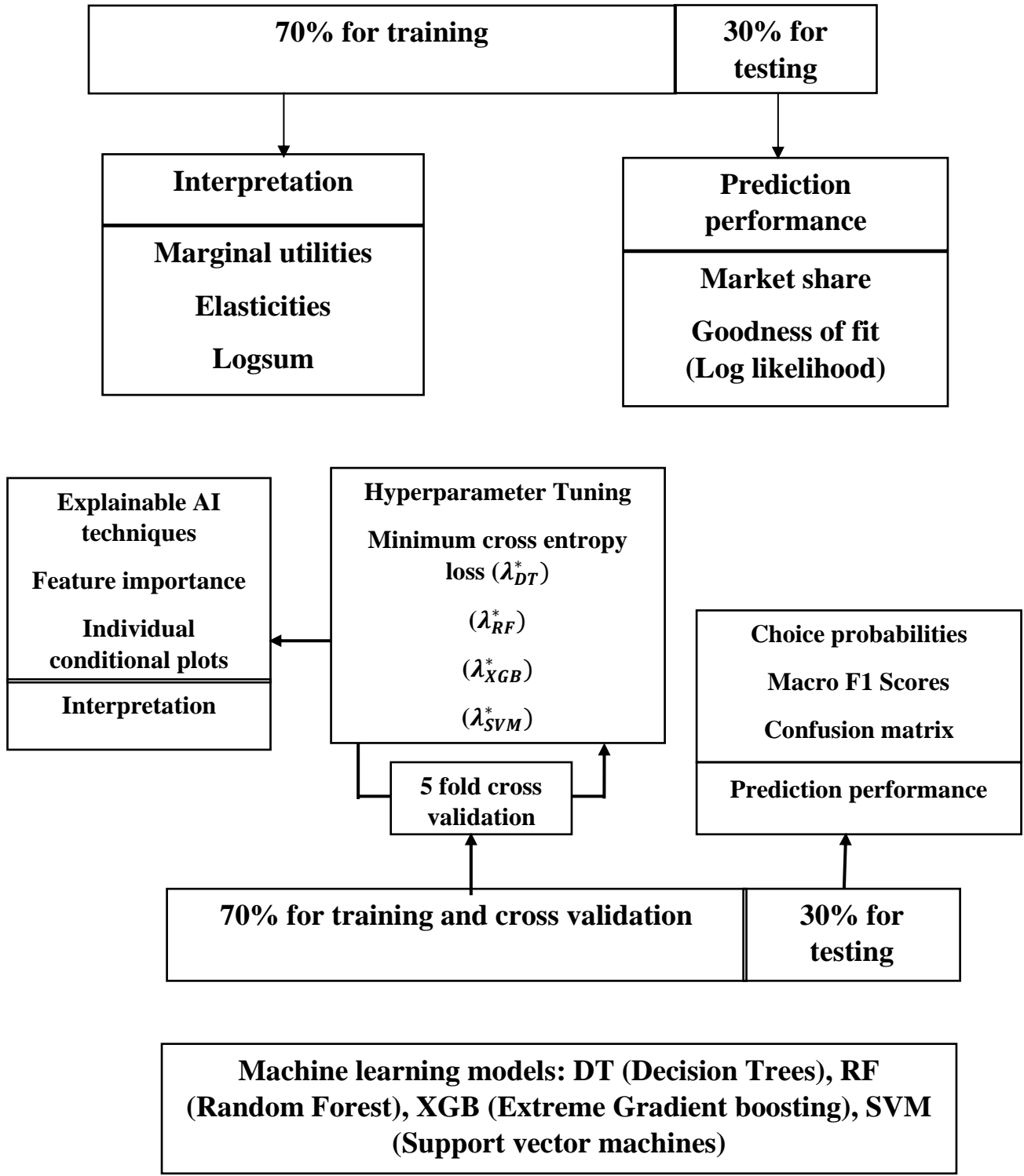

**Figure 2: Comparative framework between Multinomial logit model and Machine learning models**

**4.1 Multinomial Logit Model (MNL)**

Discrete choice modelling-based framework of MNL is used to understand and predict the mode choice behaviour of individuals (Ben-Akiva and Lerman, 1985). The model's objective is to see the impact of various attributes (travel-related and socio-economic) on the mode choice decision of individual commuters. The foundation of discrete choice theory is rooted in the economic theory of utility maximisation. The utility function for an individual $n$ choosing an alternative $i$ can be expressed as the sum of the deterministic and random components.

$$U_{ni} = V_{ni} + \varepsilon_{ni} \qquad\qquad 1$$

where

$U_{ni}$ is the total utility obtained from choosing an alternative $i$

$V_{ni}$ is the deterministic part of the utility from choosing an alternative $i$

$\varepsilon_{ni}$ is the random component

An individual $n$ will choose an alternative $i$ if the utility derived from the alternative $i$ is greater than all other alternatives $(1, 2, \ldots . j)$ in the choice set. This can be written using the following equation.

$$U_{ni} > U_{nj}; j \neq i \qquad\qquad 2$$

And the expression for the probability of choosing an alternative $i$ from the set of $J$ alternatives is:

$$P_{in} = \frac{e^{V_{in}}}{\sum_{j=1}^{J} e^{V_{jn}}} \qquad\qquad 3$$

The utility specifications which define the interactions between the dataset attributes and the observed utility ensure a high level of interpretability and model robustness. The utility specifications ensure that the model is consistent and does not contradict with the established economic theory on which the model is based.

For estimating the parameters of the MNL model, the variables used can be divided into following categories.

1. Travel related variables: travel time ($tt_i$), travel cost ($tc_i$) where 'i' stands for the class label/mode alternative.

2. Economic variables: like Household income ($HH_{inc}$) and Number of two-wheelers ($Num_{veh}$) available in the household.

3. Social variables: like Gender (which is a dummy variable with male = 0 and female = 1) and Age of the individual commuters present in the dataset.

4. Land use variables like Population density (PD) and employment density (ED) of the destination wards where individuals are commuting for different purposes.

The deterministic utility of the individual commuters in the dataset for the eight mode choices is given in the following set of equations.

$$V_i = ASC_i + \beta_{tt} \times tt_i + \beta_{tc} \times tc_i + \beta_{veh,i} \times Num_{veh} + \beta_{gen,i} \times Gender + \beta_{age,i} \times Age$$
$$+ \beta_{inc,i} \times HH_{inc} + \beta_{pop,i} \times PD + \beta_{emp,i} \times ED \qquad 4$$

where,

$ASC_i$ = Alternate specific constant for the alternatives $i$ available in the choice set

$\beta_{tt}$ = Coefficient of travel time

$\beta_{tc}$ = Coefficient of travel cost

$\beta_{veh,i}$ = Coefficient specific to the number of two- wheelers in the household

$\beta_{gen,i}$ = Coefficient specific to the gender of the individuals in the dataset

$\beta_{age,i}$ = Coefficient specific to the age of the individuals in the dataset

$\beta_{inc,i}$ = Coefficient specific to the income of the household in the dataset

$\beta_{pop,i}$ = Coefficient specific to the population density of the destination wards

$\beta_{emp,i}$ = Coefficient specific to the employment density of the destination wards

## 4.2 Machine learning models

Machine learning (ML) classifiers (such as decision trees, random forest, neural networks, gradient boosting, etc.) have proved to be powerful prediction tools to forecast phenomena related to nature, society and economics (Hernandez et al., 2016, Sahin, 2020, Kumari and Toshniwal, 2021, Chen et al., 2022, Ullah et al., 2022). Due to the non-linear nature of their prediction process, machine learning algorithms are able to capture complex relationships between the target variables and the predictors/ explanatory variables (Obulesu et al., 2018). With the advancement in modern computing and availability of huge amounts of large-scale data, they have become computationally feasible and yield state-of-the-art performance. Travel behaviour is a complex phenomenon due to the nonlinear nature of the predictors and target variables. As such, machine learning algorithms have been exploited to improve the capability of travel behaviour prediction process. However, till recently, machine learning algorithms are thought of as black box tool and therefore, the explanation and interpretation behind the results of the machine learning algorithm were largely elusive. But with recent advancements in the research [Adadi and Berrada, 2018] to improve the interpretation and rationale behind the ML models' decision output, it has become easier to summarise the behaviour of a model and explain the role played by a predictor in the final decision which further enables policy makers to identify key variables for effective policy making. A brief overview of machine learning classifiers used in this study is described in the following section.

### 4.2.1 Decision Trees

A decision tree (DT) is a supervised learning classifier which can handle extensive datasets (Sarker, 2020). It uses decision and leaf nodes: the former makes decisions with multiple branches, while the latter terminates the tree as outputs. This method iteratively dichotomizes features based on either maximum information gain or minimum entropy.

### 4.2.2 Random Forests

Random forests (RF) (Sarker, 2020) is an ensemble machine-learning algorithm combining multiple decision trees through bootstrapping and random feature selection. It harnesses various training samples randomly chosen from the main dataset with replacements. This ensemble approach often yields predictions more accurate than individual machine learning models.

### 4.2.3 Extreme Gradient Boosting

Extreme Gradient Boosting (XGBoost) is an ensemble technique utilizing decision trees for both regression and classification. Proposed by Friedman in 2001 (Sarker, 2020), it builds an ensemble through an iterative approach, focusing on estimating the residual errors of the existing model. XGBoost combines insights from multiple trees to enhance the overall model's accuracy.

### 4.2.4 Support Vector Machines (SVM)

SVM, a widely used ML technique, aims to create an optimal hyperplane that divides the n-dimensional feature space into distinct classes (Sarker, 2020). This is achieved by transforming input data into a higher-dimensional space using a kernel function, ensuring classes are linearly separable.

### 3.3 Discrete choice models vs machine learning models

Within the context of statistical methods, the emphasis is on inference and interpretability (Tukey, 1977; Letham, 2015). Statistical models are designed to make inferences about the population based using sample data, which involves hypothesis testing, parameter estimation, and uncertainty quantification (Gelman et al., 2013). These methods rely on assumptions about the underlying data distribution and offer interpretable coefficients or parameters that provide insights into the relationships between variables (Agresti, 2002). Statistical significance is assessed through hypothesis testing to determine whether the likelihood of observed effects is due to chance or, it reflects true associations (Fisher, 1925).

On the other hand, machine learning methods prioritize prediction accuracy and scalability (Bishop, 2006, Avesani et al., 2005). These methods aim to build models that automatically learn patterns and make accurate predictions or decisions on new data (Hastie et al., 2009). Machine learning algorithms leverage large datasets for training and focus on capturing complex patterns, but often sacrifice interpretability (Ribeiro et al., 2016) and are prone to overfitting and lack of generalizability. Neural networks and ensemble methods, for example, operate as black boxes, making it challenging to understand their inner workings (Montavon et al., 2018). Another challenge with these techniques is the need for huge amounts of data and heavy computing resources.

Scalability is another key aspect of machine learning, enabling efficient processing of large and complex datasets (Al jarrah et al., 2016). This scalability makes machine learning methods well-suited for big data applications where statistical methods may face computational limitations (Qiu et al., 2014, Zhou et al., 2017).

In summary, statistical methods prioritize inference and interpretability, providing insights into relationships and generalising about the population. On the other hand, machine learning methods prioritize prediction accuracy and scalability, capturing complex patterns but sacrificing interpretability. The choice between these approaches depends on the specific goals and requirements of the analysis, balancing the trade-off between interpretability and prediction performance ( refer table 3).

**Table 3. Comparison between machine learning and discrete choice model approach.**

|   | Discrete choice model | Machine learning model |
|---|---|---|
| 1 | Problem definition and formulation | Problem definition and formulation |
| 2 | Data collection and preprocessing | Data collection, preprocessing and partition |
| 3 | Selection of a closed form equation | Architecture and learning type selection |
| 4 | Parameter estimation | Hyperparameter optimisation for error minimisation |
| 5 | Testing using goodness of fit | Testing on new data and comparison with other models using F1 score |

From a behavioral analysis standpoint, Discrete Choice Models (DCMs) offer the significant advantage of interpretability, allowing for direct computation of key behavioral indicators such as Willingness to Pay (WTP) using regression coefficients (Train, 2009). This makes DCMs more appealing than Machine Learning (ML) models, which are often regarded as black-box approaches. However, DCMs require *a priori* assumptions about the underlying functional form of the utility function and the distribution of unobserved heterogeneity, which can result in model misspecification and reduced predictive power (Yahia et al., 2023). On the other hand, ML techniques relax the distributional assumptions of the model, enabling the estimation of choice probabilities without explicitly specifying the utility function.

A comprehensive comparison study by Zhao et al. (2020) explores logit models and ML methods, encompassing both prediction and behavioral analysis. The findings reveal that Random Forest (RF) exhibits the highest accuracy, although caution is advised when computing marginal effects and elasticity with standard RF due to potential unreasonable behavioral outcomes. Hensher and Ton (2000) compare artificial Deep Neural Networks (DNNs) with the Nested Logit Model (NL) in the context of commuting mode choice. While NL demonstrates superior accuracy in predicting aggregate market shares, DNNs outperform NL in predicting individual choices at a disaggregate level. Hence, no clear superiority is evident between the two approaches. Wang et al. (2020) and Ramsey and Bergtold (2021) demonstrate that DNNs can yield economic indicators similar to those derived from classical discrete choice models. These studies interpret the hidden layers of DNNs as utility functions, enabling the computation of marginal effects and WTP measures.

Wang et al. (2020) identify three challenges associated with DNNs, namely sensitivity to hyperparameters, model non-identification, and local irregularity. These challenges involve balancing the trade-off between approximation and estimation errors, optimizing the training process to identify the global optimum, and addressing locally irregular patterns in estimated functions. These challenges can lead to unreliable economic information, particularly when the sample size is small. Wang et al. (2020) and Ramsey and Bergtold (2021) highlight the difficulty of achieving stable and accurate DNNs. Unstable training algorithms can cause significant changes in estimation results with minor variations in the training dataset. To address this, the authors propose a bootstrapping DNN framework that involves resampling with replacement to obtain multiple parameter estimates.

## 4.4 Explainable AI

The lack of transparency and interpretability in AI systems, often referred to as the "black box" problem, poses a significant obstacle to understanding the decision-making process (Miller, 2019). AI models operate opaquely, making it challenging for humans to comprehend the rationale behind their outputs. This lack of transparency raises concerns regarding accountability, fairness, and the potential risks associated with AI decision-making. To tackle this challenge, researchers have introduced the field of Explainable AI (XAI) [Holzinger et al., 2020], which aims to develop methods and techniques to provide interpretable and understandable explanations for AI system decisions and actions.

Explainable AI strives to overcome the limitations of the black box problem through various approaches, including rule-based methods, model-agnostic techniques, and interpretable machine learning models (Guidotti et al., 2018). Rule-based methods use logical rules to explicitly represent the decision-making process, promoting transparency and clarity in explanations. Model-agnostic techniques, such as Local interpretable model-agnostic explanations (LIME) and SHapley Additive exPlanations (SHAP) [Ribiero et al., 2016; Lundberg and Lee, 2017], approximate the underlying model's behavior to provide explanations after the fact. Interpretable machine learning models, such as decision trees and linear models, inherently possess transparency and can offer intuitive explanations for their predictions.

The evaluation of AI models' explanations and user perception play critical roles in the adoption of XAI systems (Doshi-Velez & Kim, 2017). Assessing the quality and effectiveness of explanations is essential to ensure their comprehensibility, accuracy, and relevance to the decision-making context. Research indicates that individuals tend to trust AI systems more when explanations are provided [Samek and Müller, 2018; Arrieta et al.,2020]. Establishing evaluation metrics and benchmarks that accurately gauge the quality of explanations is crucial to drive advancements in XAI techniques.

Explainable AI finds applications in various domains, including healthcare, finance, and autonomous systems [Došilović et al., 2018; Srinivasu et al., 2020, Weber et al., 2020; Gohel et al., 2021]. In healthcare, XAI empowers clinicians by enabling them to understand the reasoning behind AI-generated recommendations, facilitating decision-making and enhancing patient trust. In finance, explainability is vital for regulatory compliance and ensuring transparency in algorithmic trading systems. For autonomous systems such as self-driving cars or drones, XAI plays a pivotal role in enabling users to comprehend the decisions made by these systems, thereby improving safety and instilling trust (Adadi and Bouhoute, 2023).

In conclusion, the development of explainable AI stems from the need to address the lack of transparency in AI systems. Through diverse techniques and approaches, XAI aims to provide interpretable explanations, fostering transparency, accountability, and trust in AI decision-making processes.

## 5. Experiments

The ML experiments are performed by splitting the dataset into train test sample in the ratio of 70 : 30. The data samples are normalised using min-max normalisation so that it lies in the range [0,1]. To fine-tune the hyperparameters, a predefined range is initialised for each of the hyperparameter. The performance of the classifiers, based on different hyperparameter combinations, is assessed using a 5-fold cross-validation approach. In this approach, the dataset is partitioned into five distinct subsets. For every hyperparameter combination, the model is trained on four of these subsets and validated on the fifth. The F1-Score, which measures the model's accuracy, is calculated using the predictions from the model and the actual values from the validation subset. This procedure is iterated for each of the five subsets, resulting in an average F1-Score for that specific hyperparameter combination. If a particular combination yields an F1-Score that surpasses the previous best, the optimal hyperparameters are updated accordingly.

### 5.2 Machine learning evaluation metrics

Accuracy is a common metric in machine learning [Ferri et al., 2009]. But when dealing with datasets that aren't balanced, using accuracy can give a distorted view of results. For this reason, this study uses the Macro F1-Score as its performance measure, which is based on the confusion matrix. The predictions made by a classifier are referred to as prediction values, while the actual or expected outcomes are known as actual values. Both of these can be classified as either Positive or Negative. A "true positive" happens when the classifier correctly assigns a positive label to an actual positive instance. A "true negative" is when the classifier correctly labels a negative instance as negative. Conversely, a "false negative" arises when a genuinely positive instance is incorrectly identified as negative, and a "false positive" occurs when a genuine negative instance is misclassified as positive. The  evaluation metrics used in this study are defined as follows:

$$Accuracy \ = \ \frac{TP + TN}{TP + TN + FP + FN} \ \times \ 100 \qquad 5$$

$$Precision \ = \ \frac{TP}{TP \ + \ FP} \qquad 6$$

$$Recall \ = \ \frac{TP}{TP \ + \ FN} \qquad 7$$

$$F1 \ score \ = \ \frac{2 \times (precision \times recall)}{precision \ + \ recall} \qquad 8$$

$$Where, TP \ = \ True \ positive$$

$$FP = False \ positive$$

$$TN = True \ negative$$

$$FN = False \ Negative$$

The accuracy of the model is estimated by F1-score which is a harmonic mean of precision and recall. F1-score ranges between 0 and 1, where 0 can be interpreted as worst score and 1 can be interpreted as best score.

## 5.3 Hyperparameter Tuning

The machine learning models have set of preconfigured parameters that control the learning process. These are called hyperparameters and depends on the algorithm type, say for example, the decision trees hyperparameters like MSL (minimum number of samples required to be a leaf node), MD (maximum depth of the tree) and CCP (cost complexity pruning). Similarly, other algorithms have their own set of hyperparameters (please see Table 4 for detailed description). Optimisation of hyperparameters is done independently for each ML model to arrive at the best possible cross-validation Macro F1-score. These optimal hyperparameter values are recorded and then used in the testing phase.

Feurer and Hutter (2019) formalised the problem of hyperparameter optimisation as follows. Consider the dataset D, which serves as the input to the algorithm, facilitating the process of learning for the algorithm. The hyperparameters of the algorithm is denoted by $\lambda$ such that $\lambda$ belongs to the feasible region of hyperparameters $\Lambda$, i.e $\lambda \in \Lambda$. The optimal hyperparameter set is obtained through following mathematical formulation.

$$\lambda^* = \arg\min_{\lambda \in \Lambda} E\left(D_{\text{train}}, D_{\text{valid}}\right) \sim \mathcal{D}_{\mathcal{V}}(A_\lambda; D_{\text{train}}, D_{\text{valid}}) \qquad 9$$

In order to approximate the expectation in Equation (), a cross-validation approach is employed. The dataset $D$ is divided into $K$ distinct subsamples, or folds, labeled as $D_1, D_2, \dots, D_K$. For each fold, denoted as $D_k$, the $A_\lambda$ classifier is trained on the complement set $D_{train}^k$ (consisting of all samples except those in $D_k$) and then validated on $D_k$ itself. This process is repeated for each fold, resulting in $K$ separate training and validation iterations.

To estimate $E(D_{train}; D_{valid}) \sim DV(A_\lambda; D_{train}; D_{valid})$, the sample mean of the performance metric $V(A_\lambda; D_{train}; D_{valid})$ across all folds is considered. In summary, this cross-validation procedure is commonly referred to as $K$-fold cross-validation.

The aim of this work is to improve the performance of each machine learning algorithm by tailoring it to the problem at hand (in this case to travel mode choice problem). The hyperparameters for all the algorithms with respect to the dataset were tuned using five-fold cross-validation technique. The process of hyperparameter tuning follows the same approach as given in the work of Sethi et al., 2022.

**Table 4. Hyperparameters and its ranges.**

| Name of hyper parameters | Description | Range |
|---|---|---|
| **Decision Tree** | | |
| MSL | The minimum number of samples required to be at a leaf node. | [1,20] |
| MD | The maximum depth of the tree. | [1,20] |
| CCP | Complexity parameter used for Minimal Cost-Complexity Pruning | |
| **Random Forest** | | |
| N | Number of estimators | [1,10,100,10 00,10000] |
| MD | The maximum depth of the tree. | [1,11] |
| ***Support vector Machines (SVM) with Radial Basis Function kernel*** | | |
| c | Regularization parameter. The strength of the regularization is inversely proportional to C. Must be strictly positive. The penalty is a squared L2 penalty. | [0.1, 100.0, 0.1] |
| ***Extreme Gradient Boost (XG Boost)*** | | |
| MD | Maximum depth of the tree | [1, 3, 5, 7, 9, 11], |
| Eta | Learning rate parameter (also known as the shrinkage or eta parameter) controls the step size at each boosting iteration. | [0.1, 0.01] |
| Gamma | Minimum loss reduction parameter | [0, 0.5, 1] |

| NT | Number of boosting rounds (trees) to be built. | [100, 150,200,250, 300,350,400, 450,500] |
| --- | --- | --- |
| Min_child_weight | Minimum sum of instance weights (hessian) required to make a further partition on a leaf node of the tree | [1, 3, 5, 7, 9, 11] |

## 6. Results and Discussion

This section presents the results of all the mode choice models used in the study. Table 5 presents the parameters estimates of the MNL model.

**Table 5. Parameter estimates of the MNL model.**

| | Parameter estimates | Standard error | t-Statistics | Behavioural interpretation |
|---|---|---|---|---|
| *Alternate specific constants (ASCs)* | | | | |
| Metro (reference alternative) | 0 | --- | --- | Alternate specific constants (ASCs) indicate baseline utility levels for various transportation modes when all other factors are assumed equal, with Metro as the reference category. The highest baseline preference is for walking (ASC: 6.652) suggesting a strong inherent utility for walking over the Metro. Other modes like Bus, Two-wheeler, and Car also demonstrate significant baseline utilities, as reflected in their ASC values. These values provide a starting point for understanding preferences, onto which the effects of other explanatory variables are added to fully model the choice behavior. |
| Bus | 3.647 | 0.231 | 15.758 | |
| SR | 3.418 | 0.269 | 12.714 | |
| Auto | 2.854 | 0.278 | 10.349 | |
| Two-wheeler | 3.412 | 0.241 | 14.139 | |
| Car | 3.636 | 0.411 | 8.841 | |
| Cycle | 1.967 | 0.466 | 4.226 | |
| Walk | 6.652 | 0.265 | 25.080 | |

| *Level of service variable (LOS)* | | | | |
|---|---|---|---|---|
| Travel time | -0.083 | 0.003 | -25.077 | The impact of travel time and travel cost on mode choice is statistically significant. The parameter estimates for the travel time suggest that for one minute increase in travel time, utility decreases by 0.083 units. Similarly, the parameter estimates for travel cost indicate that for one rupee increase in travel cost, utility decreases by 0.017 units. |
| Travel cost | -0.017 | 0.002 | -8.686 | |
| *Vehicle ownership (Two-wheeler availability)* | | | | |
| Metro (reference alternative) | 0 | ---- | --- | The positive sign of coefficient specific to auto and two-wheeler suggest that the propensity to use these modes increase with an increase in two-wheeler availability compared to metro. |
| Auto | 0.221 | 0.084 | 2.632 | |
| Two-wheeler | 0.331 | 0.031 | 10.628 | |
| *Income* | | | | |
| Metro (reference alternative) | 0 | --- | --- | The parameter estimate specific to all the mode is statistically significant and hence, is retained in the model. The negative sign of mode specific coefficient suggests that as income increases, the propensity of individuals to choose other modes decrease as compared to metro. |
| Bus | -3.030E-05 | 4.800E-06 | -6.340 | |
| SR | -1.280E-05 | 4.900E-06 | -2.630 | |
| Auto | -2.640E-05 | 5.800E-06 | -4.600 | |
| Two-wheeler | -2.370E-05 | 4.600E-06 | -5.170 | |
| Car | -1.860E-05 | 5.400E-06 | -3.430 | |
| Cycle | -2.490E-05 | 1.110E-05 | -2.230 | |
| Walk | -2.960E-05 | 5.000E-06 | -5.970 | |
| ***Gender*** | | | | |
| Metro (reference alternative) | 0 | --- | --- | The coefficient of the gender variable represents the difference in the propensity to choose a mode (relative to reference alternative, which is metro in this case) between the situation when female is the commuter and the situation when male is the commuter. The coefficient specific to bus is not statistically significant which suggests there is no difference in the preference for bus as compared to metro between male and female commuters. The negative sign of the coefficient specific to shared ride, auto, two-wheeler, four-wheeler and cycle suggests that the preference for these modes as compared to metro by female is less than male. The positive sign of the coefficient specific to walk mode suggests that the preference for walk as compared to metro is higher for female compared to male. |
| SR | -0.544 | 0.131 | -4.139 | |
| Two-wheeler | -1.435 | 0.103 | -13.980 | |
| Car | -1.107 | 0.262 | -4.220 | |
| Cycle | -1.963 | 0.540 | -3.630 | |
| Walk | 0.414 | 0.103 | 4.032 | |
| **Age** | | | | |
| Metro (reference alternative) | 0 | --- | --- | The coefficient of age variable is statistically significant specific to only Shared ride and Car. A unit increase in age decreases the utility of a shared ride by 0.028 units compared to the metro, all else being the same (ceteris paribus). Similarly, a unit increase in age increases the utility of car by 0.031 units compared to the metro alternative, all else being the same (ceteris paribus). |
| SR | -0.028 | 0.005 | -6.217 | |
| Car | 0.031 | 0.008 | 4.076 | |
| **Employment Density** | | | | |
| Metro | 0 | --- | --- | Increase in employment density decreases the preference for bus as compared to metro. However, an increase in employment density increases the preference for two-wheeler and auto as compared to metro. The negative coefficient specific to cycle mode suggests a decrease in preference for cycle as compared to |
| Bus | -8.000E-05 | 4.000E-05 | -2.162 | |

| | | | | metro as employment density of the destination places where commuters are travelling increases. |
|---|---|---|---|---|
| Auto | 2.000E-05 | 1.000E-05 | 2.032 | |
| Two-wheeler | 1.000E-05 | 1.000E-05 | 1.686 | |
| Cycle | -3.500E-04 | 1.800E-04 | -1.922 | |
| **Summary statistics** | | | | |
| Log-likelihood (constants only model) | -5172.47 | | | |
| Log-likelihood at convergence | -4399.6 | | | |
| $\rho^2$ | 0.1494 | | | |
| $adj.\rho^2$ | 0.1448 | | | |
| Number of observations | 4394 | | | |
| Training accuracy | 0.674 | | | |
| Testing accuracy | 0.616 | | | |

### 6.3 Value of Time (VOT)

The value of travel time ($VOT$) in transport economics literature can be defined as the additional cost people are willing to pay to save an additional unit of time (Athira, 2016). From the marginal utility equations, it can be calculated as the fraction of the derivative of utility ($V_i$) with respect to travel time ($tt_i$) over the derivative of utility with respect to travel cost ($tc_i$).

$$VOT = \frac{\beta_{tt}}{\beta_{tc}} \qquad\qquad 10$$

$$VOT = \frac{-0.013}{-0.012} = 1.364$$

The value of time (VOT) is calculated for different segments in the sample and the results are shown in table 6.

**Table 6. Value of travel time for different population segment**

| Population segment | Coefficient of time $\beta_{tt}$ | Coefficient of cost $\beta_{tc}$ | Value of time $\frac{\beta_{tt}}{\beta_{tc}}$ |
|---|---|---|---|
| *Occupation* | | | |
| Salaried | -0.053 | -0.014 | 3.674 |
| Wage workers | -0.095 | -0.021 | 4.548 |
| Self-employed | -0.073 | -0.013 | 5.472 |
| *Activity* | | | |
| Work | -0.083 | -0.017 | 4.910 |
| Education | -0.096 | -0.021 | 4.664 |
| *Gender* | | | |
| Female | -0.099 | -0.016 | 6.225 |
| Male | -0.066 | -0.019 | 3.561 |
| *Income* | | | |
| Low income | -0.092 | -0.023 | 3.915 |
| Medium income | -0.072 | -0.011 | 6.524 |

The VOT for different population segment is calculated as the ratio of the travel time and travel cost coefficients specific to that population segment. The travel time and travel cost coefficients for each population segment is estimated by filtering data for each population segment and then estimating the MNL model parameters on this filtered dataset. Table 6 presents VOT for the population segmented by occupation, gender, trip purpose and occupation. In the occupation category, self-employed individuals have the highest value of time which indicates that they are willing to pay more to save time spent on travel, followed by wage workers and salaried individuals. For the trip purpose category, individuals commuting for work have higher VOT compared to those commuting for the purpose of education. In the gender category, the females are willing to spend almost twice compared to the male counterpart to save one unit of time spent on travel. When categorised by income, individuals in the medium income bracket are willing to spend approximately 1.66 times compared to the individuals in the low income bracket.

## 6.4 Measuring Price sensitivity

The sensitivity to change in price is measured using elasticity and is defined as percent change in consumption of good or services if there is 1% change in its price (Litman, 2017). Elasticity can be further categorized into self and cross elasticity.

### 6.4.1   Self-elasticity

Self-elasticity ($\eta_{x_i}^{P_i}$) is defined as percent change in the probability of choosing mode "$i$" ($P_i$) if there is a 1% change in the cost ($x_i$) of mode $i$. Self-elasticity for a mode $i$ is calculated using the following formula.

$$\eta_{x_i}^{P_i} = \left(\frac{\partial P_i}{P_i}\right) / \left(\frac{\partial x_i}{x_i}\right) = (1 - P_i) \times x_i \times \beta_{tc} \qquad 11$$

If there is a fare reduction policy change which results in 10% decrease in the cost of using metro, self-elasticity of metro is calculated using the following formula.

$$\eta_{cost_{metro}}^{P_{metro}} = 10 \times (-(1 - 0.0314) \times 18.426 \times (0.011)) = 1.963$$

Probability of choosing metro under fare reduction policy will be the sum of the original metro share and the self-elasticity of the metro.

$$P_{new,metro} = P_{current,metro} + \eta^{P_{metro}}_{cost_{metro}} = 3.14 + 1.963 = 4.103$$

### 6.4.2 Cross-elasticity

Cross elasticity ($\eta^{P_j}_{x_i}$) is defined as percent change in the probability of choosing mode "$j$" ($P_j$) if there is a 1% change in the cost ($x_i$) of mode $i$. Cross-elasticity for a mode $j$ is calculated using the following formula.

$$\eta^{P_j}_{x_i} = -P_i \times x_i \times \beta_{tc} \qquad\qquad 12$$

Under the fare reduction policy of metro as described above, the cross elasticity of two-wheeler is calculated by the following formula.

$$\eta^{P_{2w}}_{cost_{metro}} = 10 \times \left(-(-0.0314) \times 5.4612 \times (-0.011)\right) = -0.123$$

Probability of choosing two-wheeler under metro fare reduction policy will be the sum of the original two-wheeler share and the cross-elasticity of the two-wheeler.

$$P_{new,2w} = P_{current,2w} + \eta^{P_{2w}}_{cost_{metro}} = 20.6 - 0.123 = 20.477$$

### 6.5 Estimation of consumer benefits using log sum measure

In the context of logit model, consumer benefits as a result of transportation related policy change can be evaluated using the logarithm of the sum of the exponents of deterministic utility values across all mode choices or commonly referred to as log sum measure (Jong et al., 2007).

The consumer surplus in economics is the utility derived from choosing a particular mode of transportation in monetary terms. It is a useful tool for cost benefit analysis and is being used for transport projects appraisals in terms of cost and benefit. The change in consumer surplus can be calculated as the difference in the average consumer surplus before and after the introduction of certain policy measures which induces a change in consumer surplus value.

$$\Delta\mathrm{E}(CS_n) = \left(\frac{1}{\alpha_n}\right)\left[ln\left(\sum_i e^{V^{(1)}n,i}\right) - ln\left(\sum_i e^{V^{(0)}n,i}\right)\right] \qquad 13$$

Where:

$\Delta\mathrm{E}(CS_n)$ = Average change in consumer surplus of individual $n$ due to policy change. Superscripts 0 and 1 refer to before and after policy change.

$\alpha_n$ is the scaling factor associated with the travel cost (time) parameter. In this study, the scaling factor is taken as unity for simplification purpose.

In this study, sensitivity analysis is carried out by introducing a change in the cost of public and private transportation modes to assess the change in the probability of choosing the particular mode and the change in utility specific to a particular mode for an individual. This is further extended to calculate change in consumer surplus of the individuals. The strength of calculating consumer surplus using logit model is their ability to be aggregated across wide range of dimensions. As such, four different dimensions (Gender, Income, Trip purpose and Occupation) have been considered to compare the change in consumer surplus due to policy changes. The change in consumer surplus specific to gender variable can be written as follows.

$$\Delta \mathrm{E}^{(Male)} = \sum_{n \in Male} \Delta \mathrm{E}(CS_n) \qquad 14$$

$$\Delta \mathrm{E}^{(Female)} = \sum_{n \in Female} \Delta \mathrm{E}(CS_n) \qquad 15$$

where,

$\Delta \mathrm{E}^{(Male)}$ = Change in consumer surplus when the gender of all the individual $n$ in the population is male

$\Delta \mathrm{E}^{(Female)}$ = Change in consumer surplus when the gender of all the individual $n$ in the population is female

The change is introduced by increasing the cost of private transportation modes (Two-wheeler and Four wheeler) by 20% and decreasing the price of metro to the price of bus for each individuals while keeping everything else fixed. The analysis of the impact of policy change on consumer surplus specific to different population segment is shown in the figures 3 to 6.

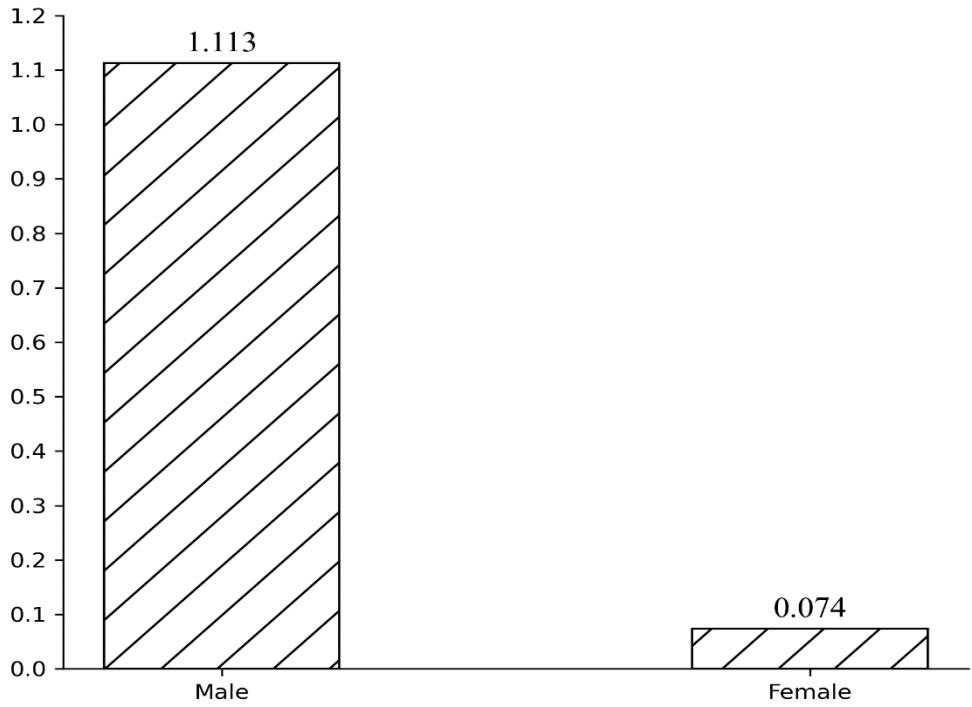

**Figure 3. Consumer surplus gender wise using logsum measure.**

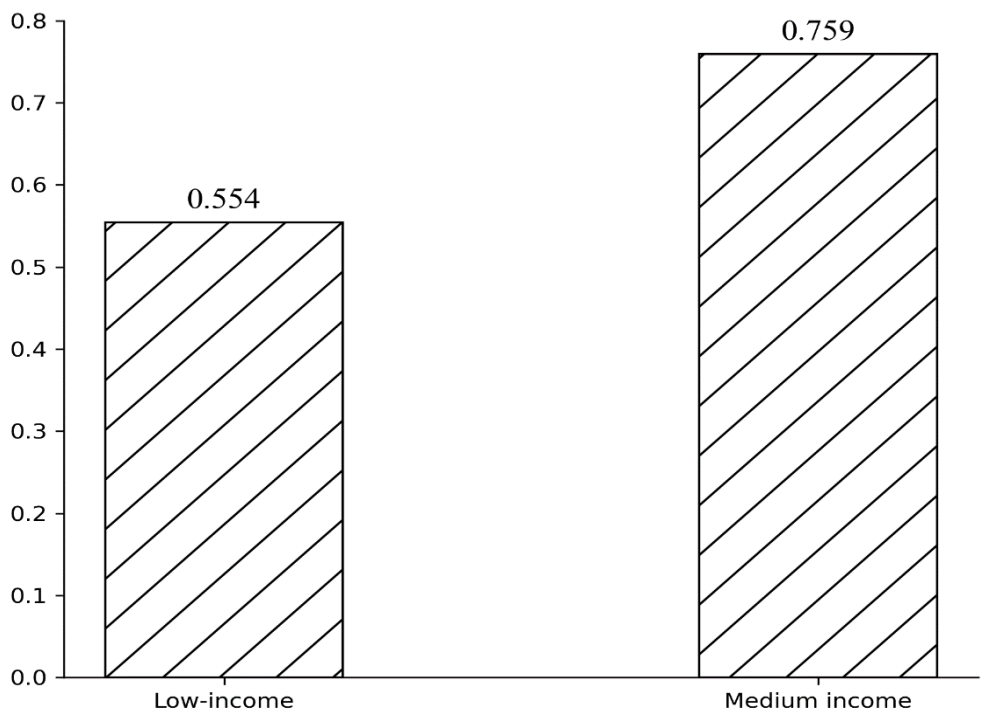

**Figure 4. Consumer surplus by income using logsum measure.**

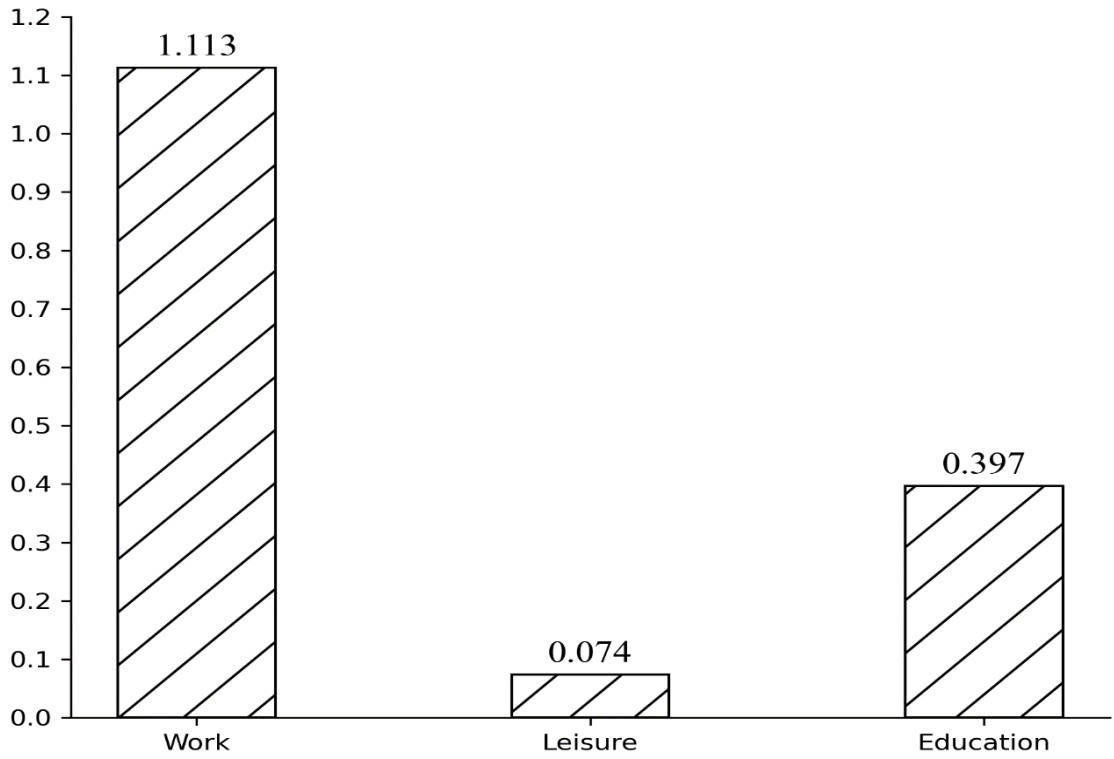

**Figure 5. Consumer surplus by trip purpose using logsum measure.**

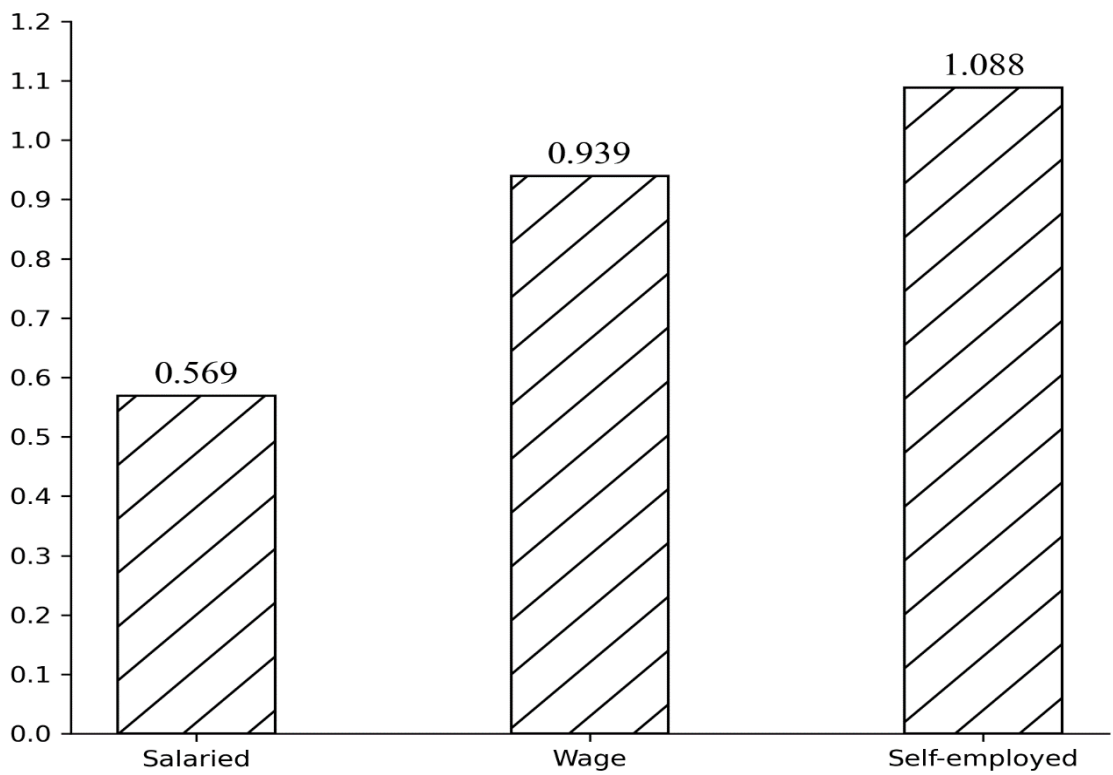

**Figure 6. Consumer surplus by occupation type using logsum measure.**

The change in consumer surplus is more pronounced for male compared to females with the given policy change. In case of trip purpose, the leisure commute is experiencing highest change in consumer surplus , followed by work and education related trips. Regarding income category, medium income individuals exhibit significantly higher change in consumer surplus compared to the low-income individuals. In the context of occupation type, the self-employed individuals register highest change in consumer surplus followed by salaried and daily wage workers. In the next subsection, we will discuss the results obtained using machine learning classifiers. Table 7 presents the results of the machine learning classifiers.

**Table 7. Predictive performance of the models used in this study.**

|  | Best F1 Score | | Optimal Hyperparameters |
|---|---|---|---|
|  | Training data | Testing data |  |
| *Decision Tree* | 0.865 | 0.401 | MSL = 1, MD = 14, CCP = 0.000823 |
| *Random Forest* | 0.788 | 0.605 | MD = 10, N = 100 |
| *Extreme gradient Boost (XGB)* | 0.915 | 0.572 | MD = 3, LR = 0.1, NT = 1000, Min_child_weight =1 Gamma = 0 |
| *Support Vector Machine (SVM)* | 0.747 | 0.543 | c = 99.7 |

In the results, the Decision Tree exhibits high accuracy on the training set, yet considerably lower accuracy on the test set, suggesting potential overfitting. The Random Forest model offers a more even performance on both training and test sets. The XGB, much like the Decision Tree, showcases pronounced accuracy during training but less so during testing. Both SVM and MNL models provide consistent performances across training and test datasets.

## 6.6 Confusion Matrix

The confusion matrix plot is a visually interactive tool that illustrates the accuracy of predictions in a classification model. It presents the true and predicted outcomes which helps in understanding the specific classification strengths and weaknesses of the algorithms. A visual representation of confusion matrix is shown in the figure 7.

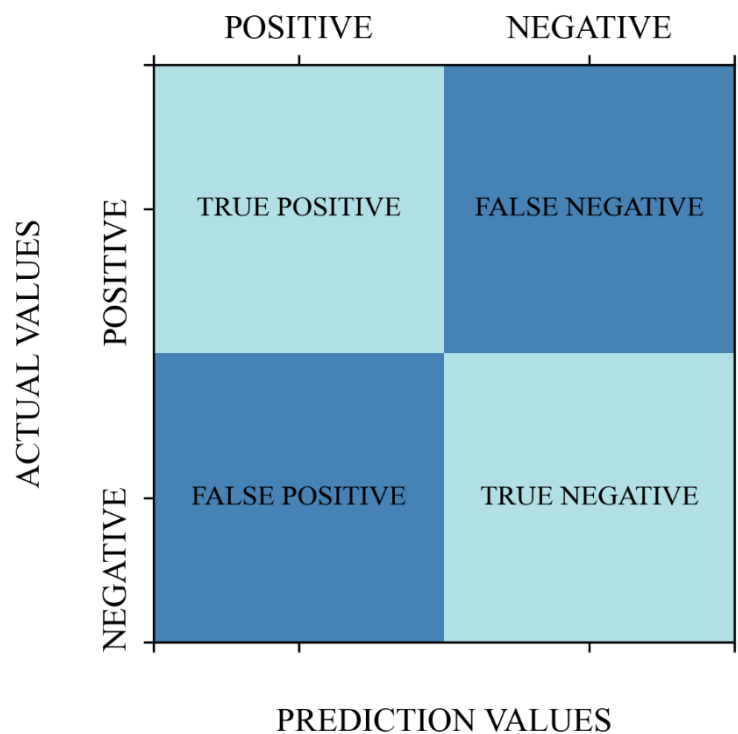

**Figure 7. Visual representation of confusion matrix.**

The confusion matrix plots for all the algorithms (which includes both the training and testing dataset) are given in figures 8 to 11.

In case of training dataset, SVM model classifies with the highest error followed by the Random Forest model. XGBoost performs with 100% accuracy on the training data, with no misclassification. It's worth checking on other datasets to ensure it's not just memorizing the answers. The Decision Tree model also shows strong accuracy percentage, with few misclassification.

However, for the test data, apart from classifying the 'walk' mode with over the 83% accuracy, decision trees perform rather sub optimally, with accuracy percentage as low as around 6% for the bus mode class. The accuracy percentage exhibited by Random Forest model on test data is very high for 'walk' (100% accuracy) and 'bike' (96% accuracy)mode class. However, the model performs poorly in classifying the 'cycle' mode, with only 10% classification accuracy. The XGBoost model precisely identifies 'walk' mode with an accuracy approaching 96%. However, the classification of 'bus' mode remains problematic, registering only 16.35% accuracy. The SVM approach yields very high accuracy of 98.05% for the 'walk' mode and an 89% for the ' two-wheeler' mode. However, its performance plummets for the 'bus' mode, evidenced by a 16.67% classification success rate. It is evident that certain modes present challenges **in** classification. The difference in the result for decision trees and XGBoost clearly shows the problem of overfitting as it is performing exceptionally well for the training data but registers suboptimal results with the test data. Unlike XGBoost and decision trees, Random forest and SVM shows consistent results with both the training and testing dataset.

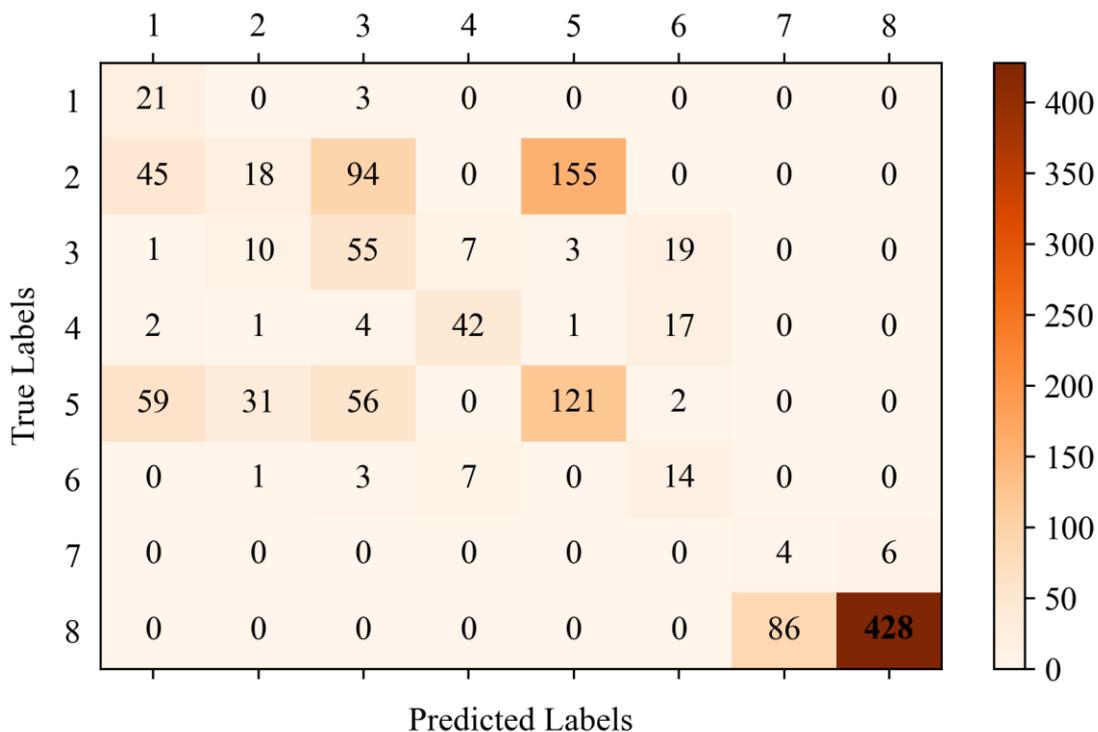

**Figure 8. Confusion matrix for decision tree model (test data).**

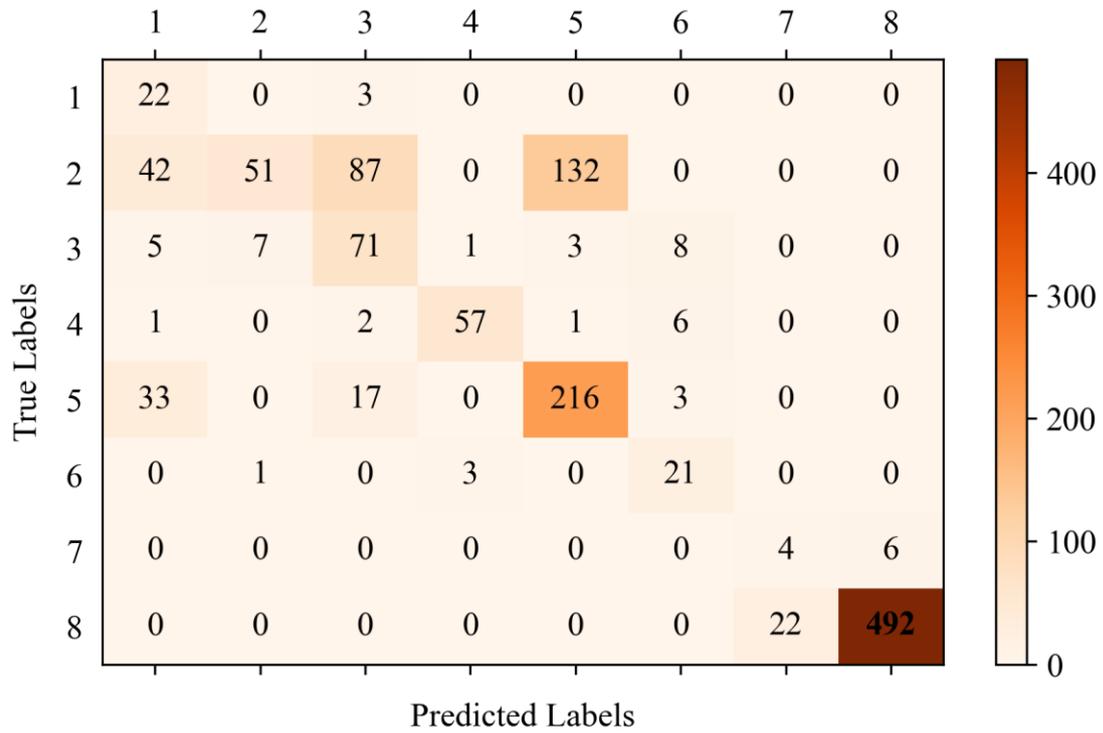

**Figure 9. Confusion matrix for XGB model (test data).**

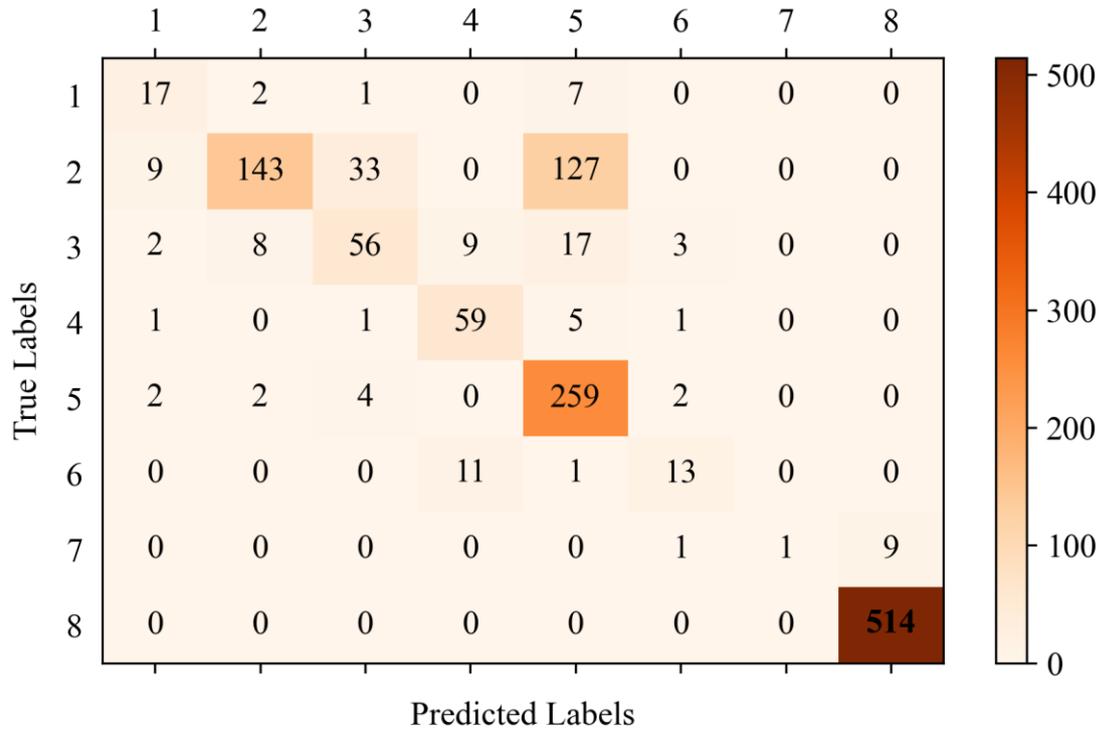

**Figure 10. Confusion matrix for RF model (test data)**.

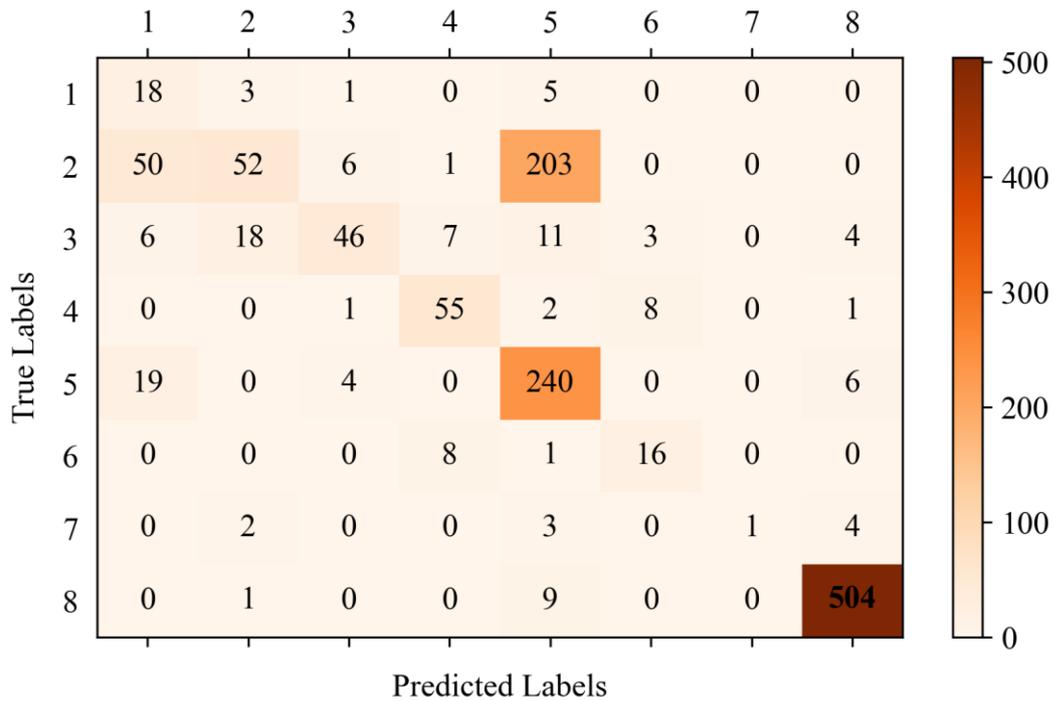

**Figure 11. Confusion matrix for SVM model (test data)**

The results of the confusion matrix on test dataset suggest that the bus mode is being classified poorly across all the models. It may be due to following reasons.

1. Overlapping Features: One of the possible reasons is that the features associated with the bus mode are not distinct enough from those of other modes like two wheeler or shared rides. If many attributes are shared between multiple modes, it might be difficult for the model to distinguish between them.

2. Interactions with other modes: Another possible reason can be confusing interactions between this mode and others which model is not able to disentangle.

3. Another reason can be the lack of enough data for the model to learn and make predictions.

Because of the possible overlapping between bus and other modes, most noticeably with the two wheeler, we calculated Euclidean distance between all the features across all the classes. Table 8 presents the pair wise distance matrix between all the features in the data for all the classes.

**Table 8. Pair wise distance matrix between all the features.**

|  | Metro | Bus | Shared Ride | Auto | Two-wheeler | Four-wheeler | Cycle | Walk |
|---|---|---|---|---|---|---|---|---|
| Metro | 0.000 | 14134.479 | 16704.041 | 16064.047 | 12499.242 | 12426.367 | 19137.973 | 20527.084 |
| Bus | 14134.479 | 0.000 | 11008.423 | **6868.627** | **8902.719** | 17073.098 | 8879.270 | 10653.437 |
| Shared Ride | 16704.041 | 11008.423 | 0.000 | 7908.438 | 4554.616 | 10611.175 | 7141.710 | 9294.185 |
| Auto | 16064.047 | 6868.627 | 7908.438 | 0.000 | 6451.631 | 15599.191 | 3660.929 | 4621.759 |
| Two-wheeler | 12499.242 | 8902.719 | 4554.616 | 6451.631 | 0.000 | 9259.575 | 7926.799 | 9771.470 |
| Four-wheeler | 12426.367 | 17073.098 | 10611.175 | 15599.191 | 9259.575 | 0.000 | 16742.966 | 18402.023 |
| Cycle | 19137.973 | 8879.270 | 7141.710 | 3660.929 | 7926.799 | 16742.966 | 0.000 | 2761.726 |
| Walk | 20527.084 | 10653.437 | 9294.185 | 4621.759 | 9771.470 | 18402.023 | 2761.726 | 0.000 |

Lower the distance values between two pair of classes, higher is the similarity between the features specific to those class pairs. From table 8, it is evident that the distance between bus and auto (**6868.627**) is least followed by the distance between bus and two-wheeler (**8902.719**) which can be one of the reason for misclassification of the bus class across all the machine learning algorithms.

### 6.7 Choice probabilities

The choice probabilities of an individual data instance can be calculated using different methods specific to each algorithm used in this study. In case of random forest, the choice probability for an instance ($P_i$) is derived from the fraction of trees that vote for a particular class and can be written as

$$P(y_i = c|x_i) \ = \ \frac{Number\ of\ trees\ voting\ for\ a\ particular\ class}{Total\ number\ of\ Trees} \qquad 17$$

The XGB model predicts scores for each class , which is transformed into probabilities using SoftMax function for multiclass problem. For all models, the choice prediction for an instance is simply the class with the highest predicted probability which can be written as:

$$\hat{y} = argmax_c \ P(y_i = c|x_i) \qquad 18$$

Similarly, the mode share of a class (which is the mode of transportation for this study) is the average of choice probabilities over all instances.

Modal share of class c $= \frac{1}{N} \sum_{i=1}^{N} P(y_i = c | x_i)$        19

where,

$\hat{y}$ = Predicted class label for a given instance

$P(y_i = c | x_i)$ = Probability that true class label $y_i$ of the $i^{th}$ instance is equal to class c given the feature $x_i$ of the instance

N = Total number of instances

**Table 9. Modal share calculated using different algorithms (values are in percentage)**

| | Actual share | RF | Standard deviation from actual share | XGB | Standard deviation from actual share | MNL | Standard deviation from actual share |
|---|---|---|---|---|---|---|---|
| Metro | 1.75 | 1.65 | 0.050 | 1.69 | 0.030 | 1.76 | 0.005 |
| Bus | 22.60 | 23.07 | 0.235 | 22.96 | 0.180 | 22.86 | 0.130 |
| SR | 7.04 | 7.13 | 0.045 | 7.22 | 0.090 | 7.22 | 0.090 |
| Auto | 5.28 | 5.49 | 0.105 | 5.56 | 0.140 | 5.36 | 0.040 |
| TW | 24.38 | 23.49 | 0.445 | 23.38 | 0.500 | 22.50 | 0.940 |
| Car | 2.18 | 2.35 | 0.085 | 2.37 | 0.095 | 2.09 | 0.045 |
| Cycle | 0.69 | 0.64 | 0.025 | 0.68 | 0.005 | 0.56 | 0.065 |
| Walk | 36.08 | 36.18 | 0.050 | 36.13 | 0.025 | 37.66 | 0.790 |

## 6.8 Feature importance

In XGBoost, the feature importance is calculated based on the contribution of each feature in reducing the objective function's loss during the tree construction process (Chen & Guestrin, 2016). The objective function typically used is the sum of the loss function and a regularization term. The loss function measures the discrepancy between the predicted values and the true values of the target variable.

During each iteration of XGBoost, a decision tree is added to the ensemble. The feature importance is calculated by summing up the gains achieved by each feature when splitting the data. The gain is the improvement in the objective function's loss after the split is made. Mathematically, the feature importance ($FI$) for a particular feature ($f$) is computed as:

$$FI = \sum \left( \frac{\Delta f}{\Sigma_J \Delta f_j} \right) \qquad\qquad 20$$

where $\Delta f$ represents the gain achieved by feature $f$, and $\Delta f_j$ represents the total gain across all features $j$.

In Random Forest, the feature importance is determined by measuring the average decrease in impurity caused by splitting on each feature (Breiman, 2001). The impurity can be quantified using metrics such as Gini impurity or entropy. These metrics assess the homogeneity of the target variable within each subset created by the splits.

The feature importance is computed by averaging the impurity decrease over all decision trees in the Random Forest. Mathematically, the feature importance ($FI$) for a particular feature ($f$) is calculated as:

$$FI = \sum \frac{-\Delta f}{N_{trees}} \qquad\qquad 21$$

where $-\Delta f$ represents the average decrease in impurity achieved by splitting on feature $f$, and $N_{trees}$ is the total number of decision trees in the Random Forest.

In Decision Trees, features are selected by splitting based on their ability to segregate the target variable into pure or near-pure subsets, thus reducing impurity. This impurity is measured using metric like Shannon's entropy or Gini impurity. Impurity is a measure of the disorder of the target variable in a subset. Common metrics include Gini impurity and entropy for classification tasks. Feature Importance (FI) for a particular feature $f$ is typically calculated as:

$$FI(f) = \sum_{nodes}(w_{node} \times \Delta impurity_{node})\hspace{4cm}22$$

where,

$w_{node}$ is the proportion of samples reaching the node, and $\Delta impurity_{node}$ is the impurity reduction achieved by the split made on feature f at that node. The sum is taken over all nodes in the tree where feature $f$ is used for splitting. If a feature often appears high in the tree, making important splits that segregate the data, it will have a higher importance score.

SVMs, especially those with linear kernels, find a hyperplane in the feature space that best separates the different classes. For non-linear separation, SVMs transform the feature space using a kernel trick. In SVMs, hyperplane acts as the decision boundary which is a subspace in the feature space which is one dimension lower than the feature space. In linear SVMs, the importance of a feature corresponds to the absolute value of its weight. Features with higher absolute weights are more influential in deciding the hyperplane.

Feature Importance ($FI$) for a particular feature $f$ in linear SVMs:

$$FI(f) =\mid w_f \mid\hspace{5cm}23$$

where,

$w_f$ is the weight associated with feature $f$.

In case of SVMs with non-linear kernels (like Radial Basis Functions), extracting feature importance is more complex and techniques like permutation importance is used to calculate the importance of features for non-linear SVMs (Smolinska et al., 2012).

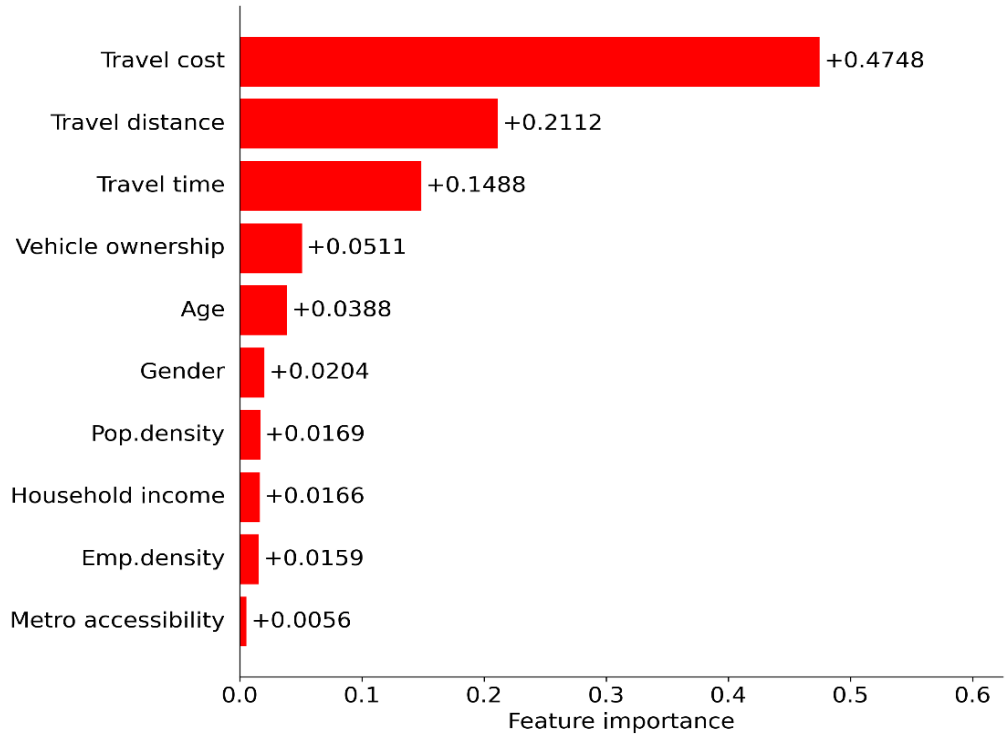

**Figure 12. Feature importance by XGBoost model**

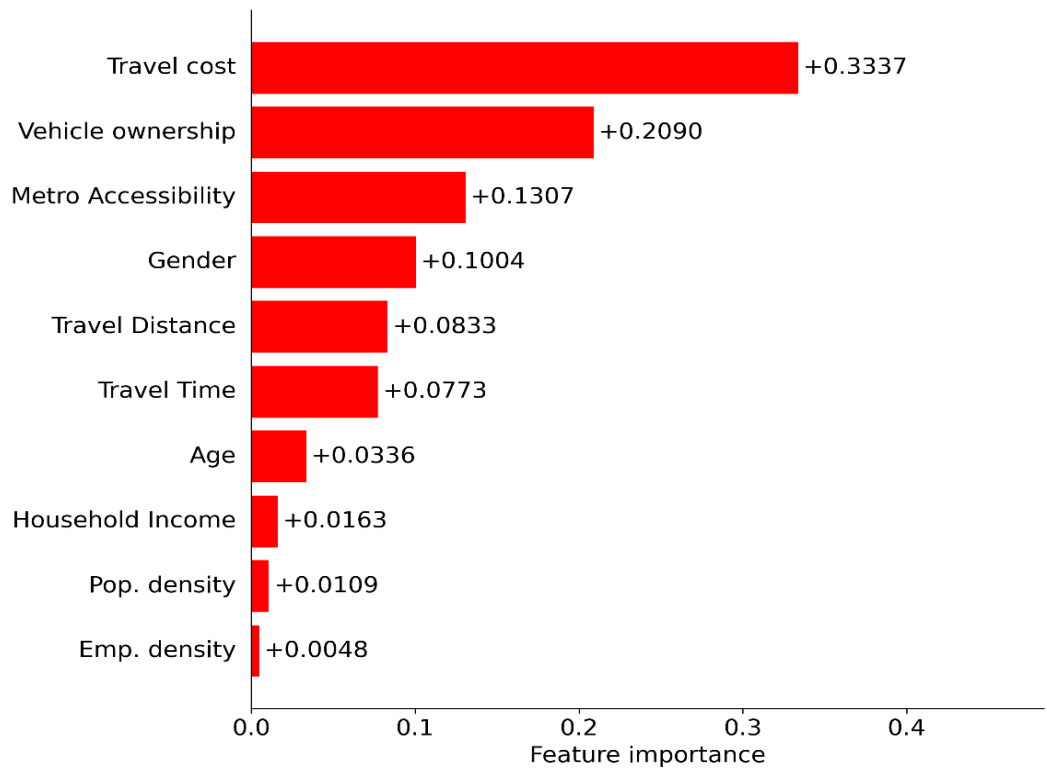

**Figure 13. Feature importance by Random Forest model**

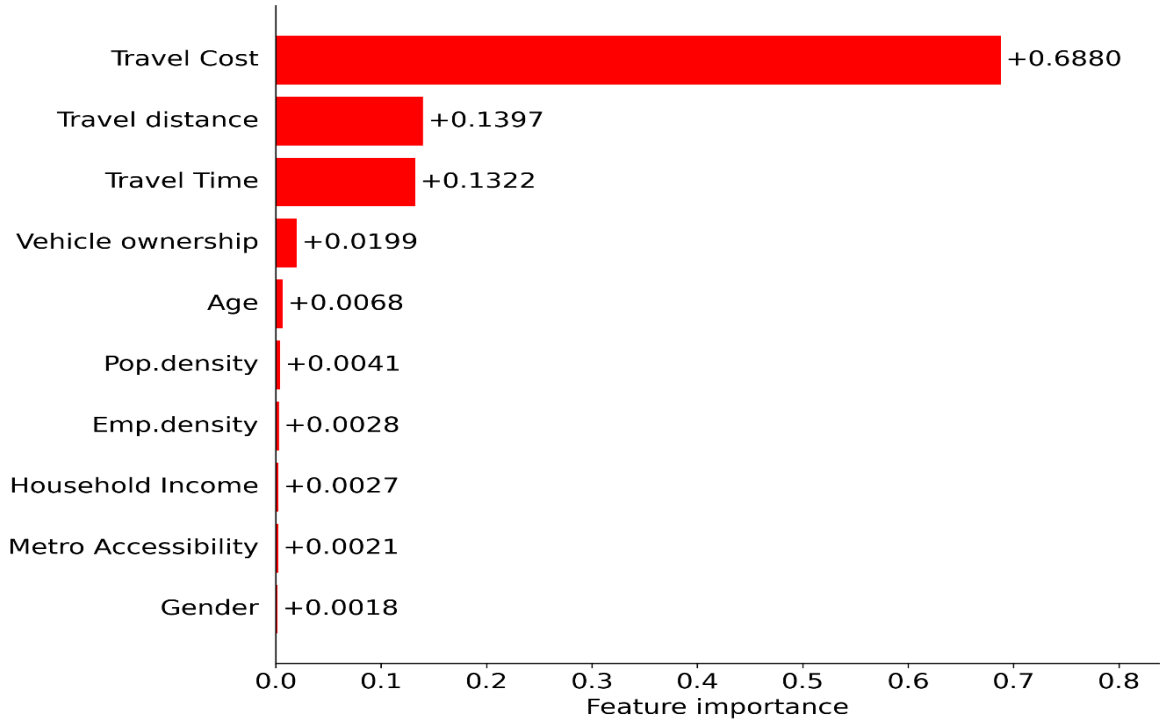

**Figure 14. Feature importance by Decision tree model.**

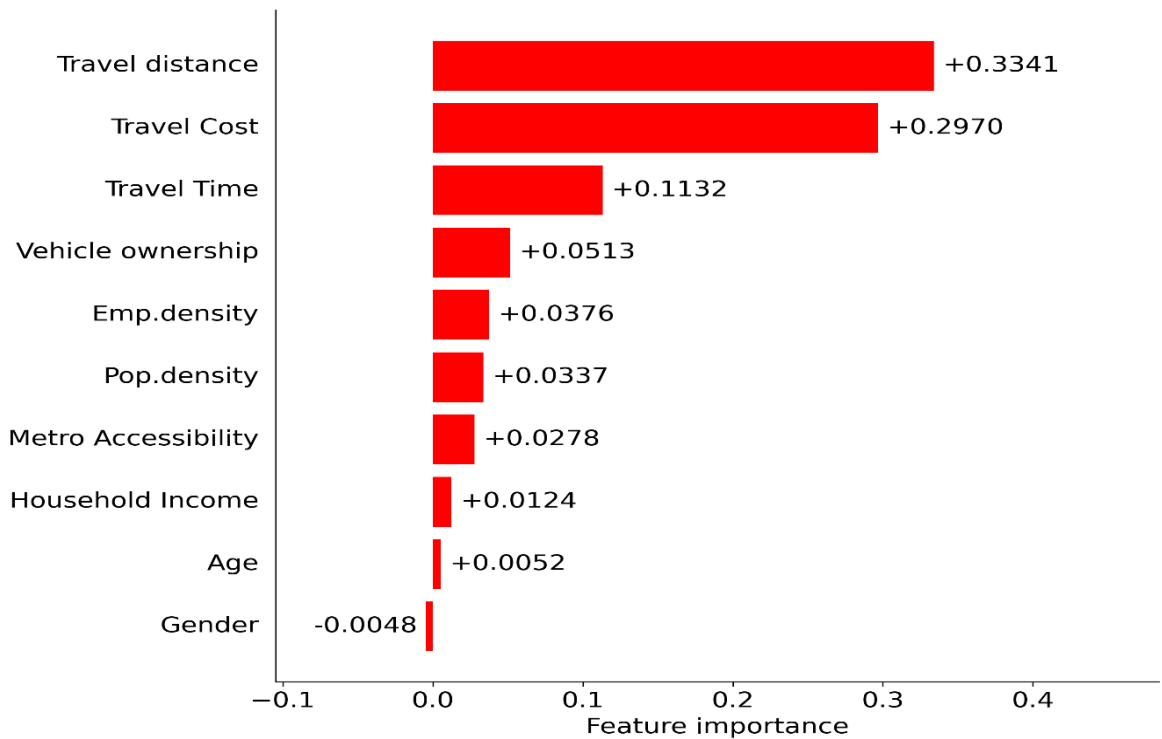

**Figure 15. Feature importance by SVM model**

Feature importance provides valuable insight into the influence of specific variables in explaining the mode choice behaviour of the individuals across the dataset. When compared across multiple algorithms, the trends or deviations in feature importance can help understand the underlying patterns or biases in each method. However, the insights derived from feature importance doesn't necessarily imply causality.

The feature importance plots for four different algorithms are shown in figures 12 to 15. Travel cost is the most dominant feature across all the models expect SVM where travel distance is the most important feature. The results suggest that travel related characteristics like cost, time and distance play the most important role in explaining the mode choice for the individuals belonging to low and middle income individuals in BBMP, followed by economic variables like the number of vehicles owned by individuals. Apart from that, land-use variables like population and employment density play a minor role in explaining mode choice. Variables like Household income, Age and Gender of the commuting individuals have the least feature importance across all the analytical models.

### 6.9 Individual Conditional Expectation (ICE)

Individual Conditional Expectation (ICE) plots (see figures 16 to 21) are localised analytical method which depict the relationship between a feature and the prediction for individual data points. While global methods like Partial Dependence Plots (PDP) show the average effect of a feature on prediction of the model, ICE plots show this effect for individual observations, thus providing insight into the model's behaviour at a granular level of single datapoint (Goldstein et al., 2015). This can be useful for understanding interactions and heterogeneity in the effects of features.

The mathematical expression for an ICE plot can be formulated as below.

Let $f$ be the predictive class of model trained on a given dataset. $X$ is the set of features and $X_{,i}$ denotes the $ith$ feature for which ICE plot is being constructed. For a single instance $n$ with its complete set of feature attribute $X_{n,}$, the ICE curve for feature $i$ is defined as:

$$ICE_{X_{n,i}} = f(X_{n,}|X_{n,i} = t) \qquad 24$$

Where:

$ICE_{X_{n,i}}$ = ICE plot for the $n^{th}$ instance when the feature $X_{n,i}$ is experimentally set to various values of $t$

$f(X_{n,}|X_{n,i} = t)$ represents the prediction of model $f$ for the $n^{th}$ instance when $X_{n,i}$ is set to t and all other features in $X_{n,}$ is set to their original value. The ICE plot is generated by varying the value of $t$ over the range of observed values for $X_{,i}$ (entire range of the $ith$ feature across all instances in the dataset).

The ICE values predict the probabilities of choosing a particular class for each instances in the dataset as the value of our feature of interest changes, keeping all other feature values constant. Thus, each lines in the ICE plot corresponds to individual data instances and traces the change in model's predicted probabilities for the chosen class with change in the feature values. In this study, ICE values have been plotted by sampling 50 evenly spaced data points from the test set across a range of feature values. This strategy enhances the visualisation and interpretability along with computation efficiency and helps in deriving meaningful insights from the ICE plots.

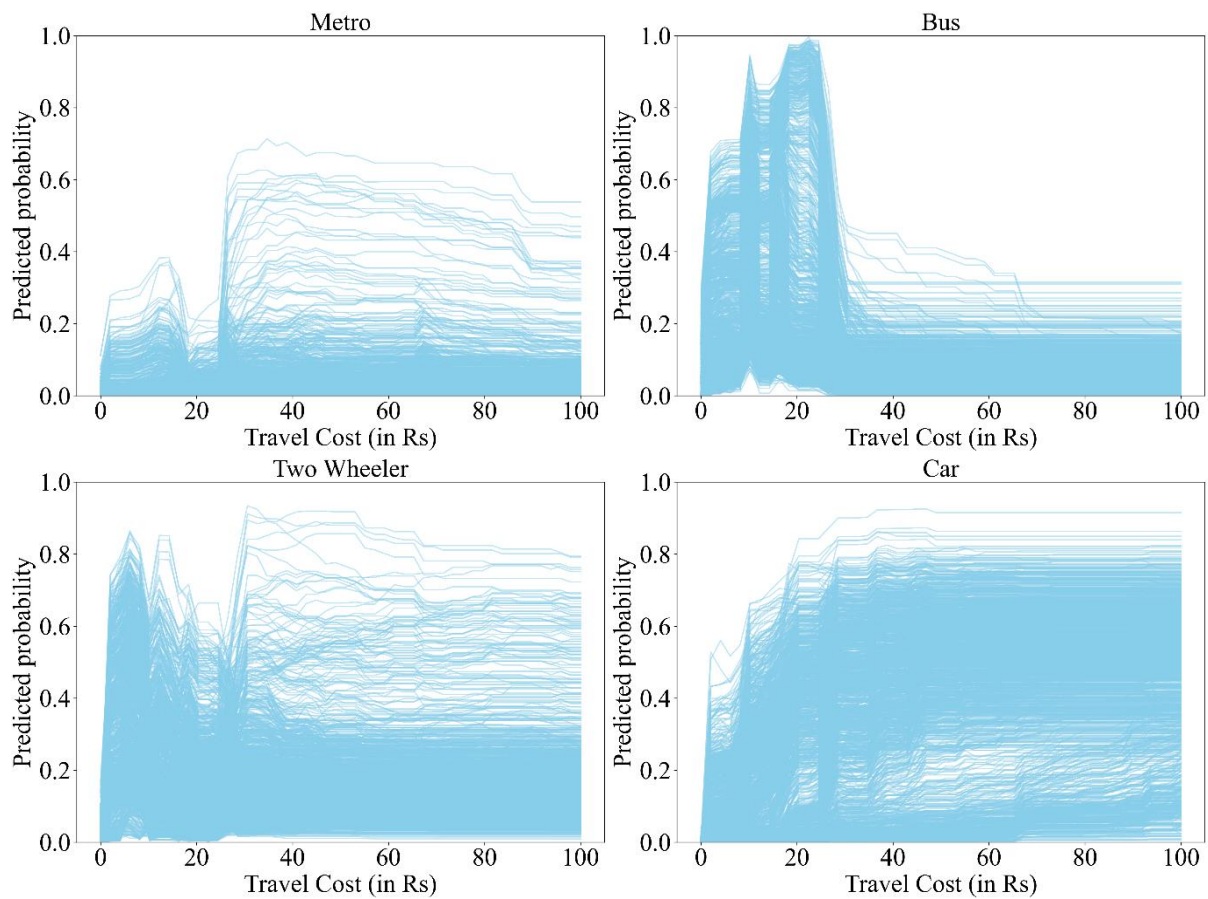

**Figure 16. ICE plots for Random Forest model between travel cost and predicted probabilities for metro, bus, two-wheeler and car.**

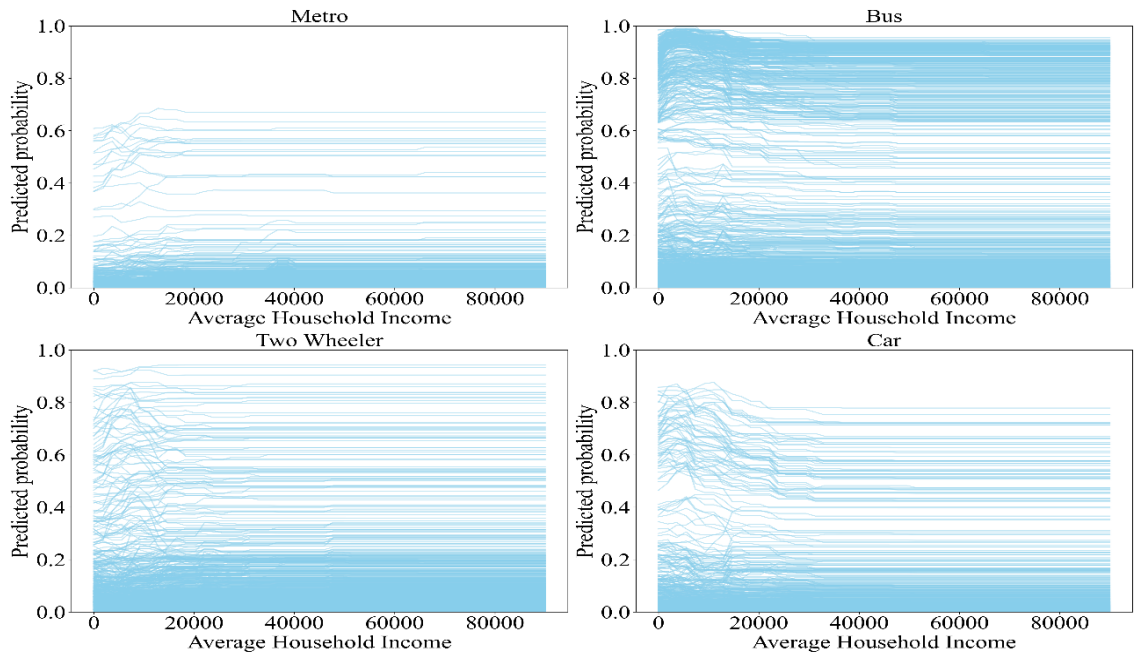

**Figure 17. ICE plots for Random Forest model between average household income and predicted probabilities for metro, bus, two-wheeler and car.**

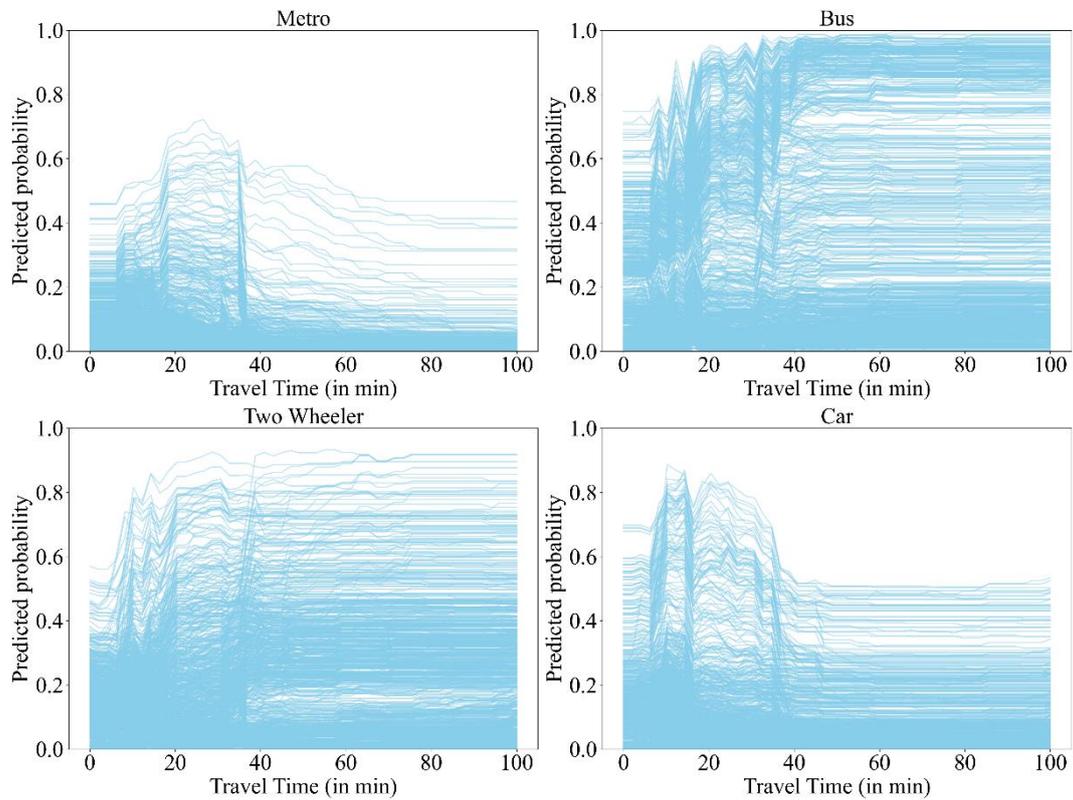

**Figure 18. ICE plots for Random Forest model between travel time and predicted probabilities for metro, bus, two-wheeler and car.**

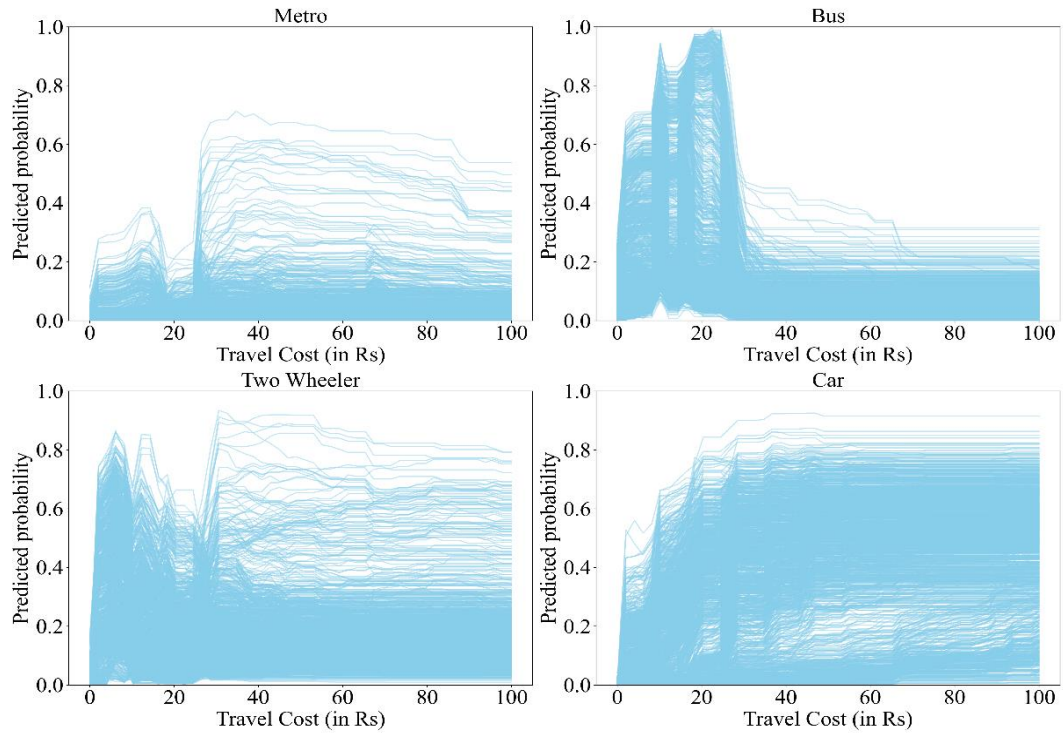

**Figure 19. ICE plots for Extreme gradient boost model between travel cost and predicted probabilities for metro, bus, two-wheeler and car.**

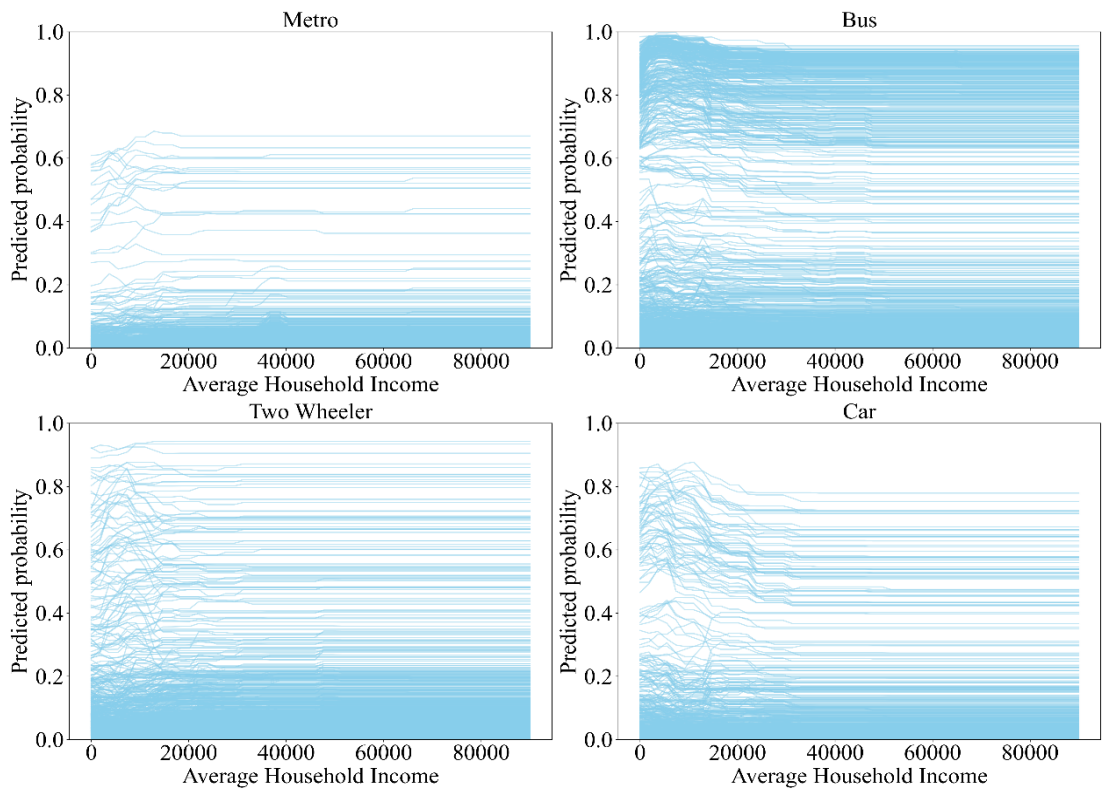

**Figure 20. ICE plots for Extreme gradient boost model between average household income and predicted probabilities for metro, bus, two-wheeler and car.**

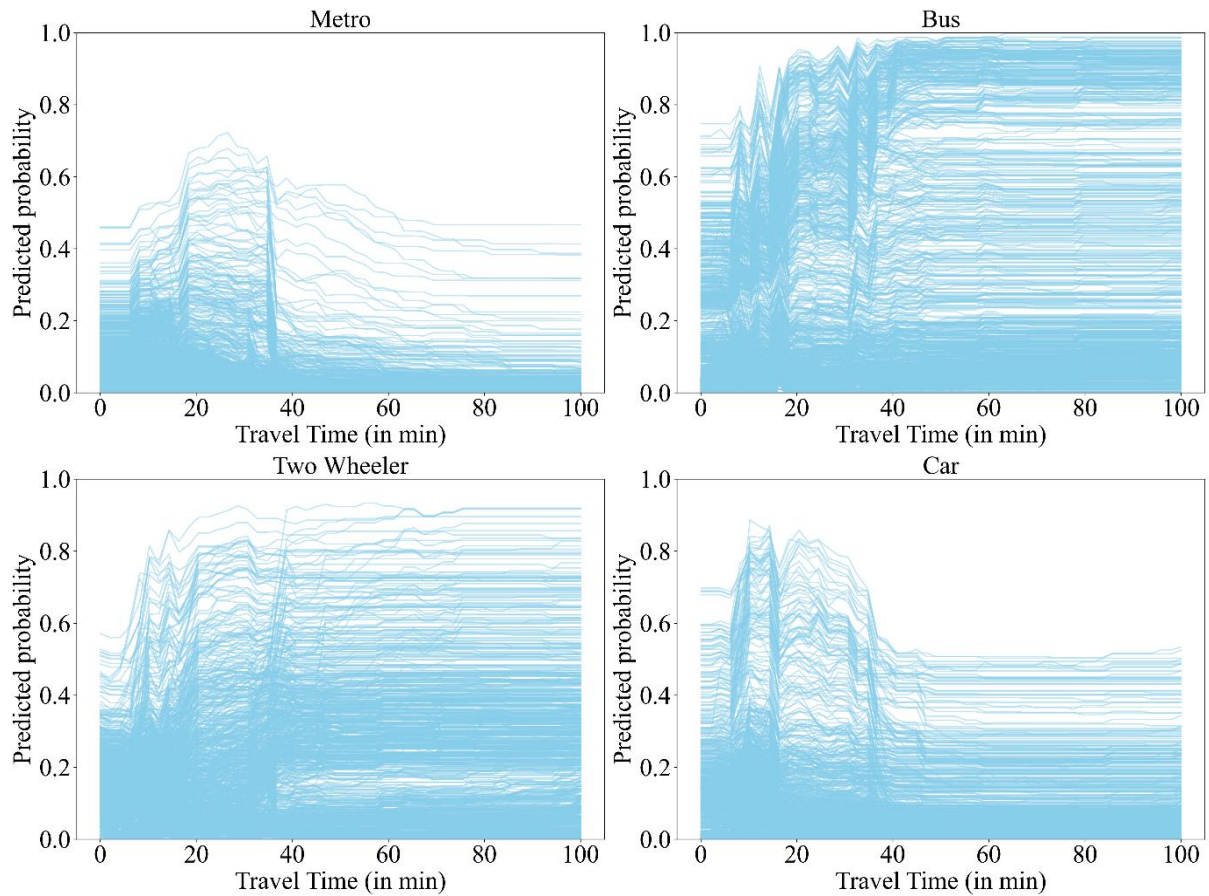

**Figure 21. ICE plots for Extreme gradient boost model between travel time and predicted probabilities for metro, bus, two-wheeler and car.**

Each lines in the ICE plot represent how change in the feature values impact the predicted probabilities of choosing a particular mode of transportation for a single instance.

As shown in figure 16 (RF) and 19 (XGB), the probability of choosing metro is higher at lower cost and drops sharply as cost increases. However, for some observations, choice of metro is less sensitive to increase in the cost. The probability of choosing bus is high when the cost is low which is being reflected in both the models as there is a sharp decrease in the probability with the travel cost exceeding INR 30. Probability of choosing two wheeler is consistently high across different observations suggesting the higher preference for two wheelers. A high probability of choosing car across all cost ranges exist suggesting that for a segment of the population, cost is not a deciding factor in their mode choice decision.

There is an optimal time range till when the passengers are willing to use metro. Probability decreases with increase in travel time for metro. Preference for bus at lower travel time is less which increases with the increase in travel time and then saturates which suggests that passengers are willing to use bus for longer travel time. There is preference for using two wheeler for medium travel time range. Travel time has less impact on the decision to use two wheeler. Probability of using car increases with travel time till a certain point (around 40-50 minutes) beyond which it decreases.

Further, individual predictions of the ICE can be averaged out to assess the broader impact of variability in feature values on model's predictions. This method allowed us to create different set of scenarios by varying the value of different features and test the sensitivity of model's outcome to such changes. In this study, five different scenarios are created for each features – 10% and 20% increase in travel cost, 10% increase in average household income and 10% and 20% decrease in travel time. The modified dataset with different individual scenarios are used as input dataset for the trained model (Random forest and XGBoost) to extract the probabilities of choosing a particular mode, which is then compared for each scenario. This allows us to infer the impact of each feature on the average probabilities of choosing a particular mode of transport. Figures 22 to 27 presents the ICE plots which has been averaged out for all instances for four different modes of travel (Metro, Bus, Two-wheeler and Car). The average ICE plot shows the trend of predicted probability of each travel mode as the feature values are modified as per different scenarios.

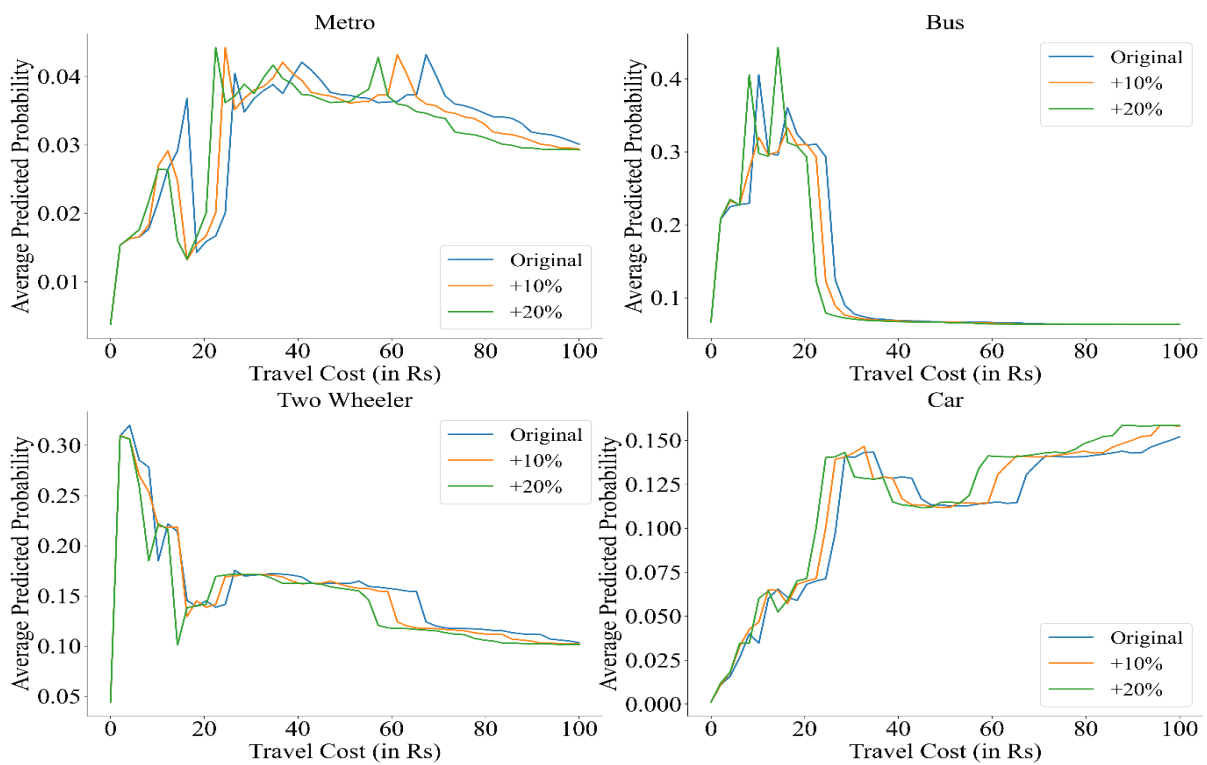

**Figure 22. ICE plots for Random Forest model between travel cost and average predicted probabilities for metro, bus, two-wheeler and car.**

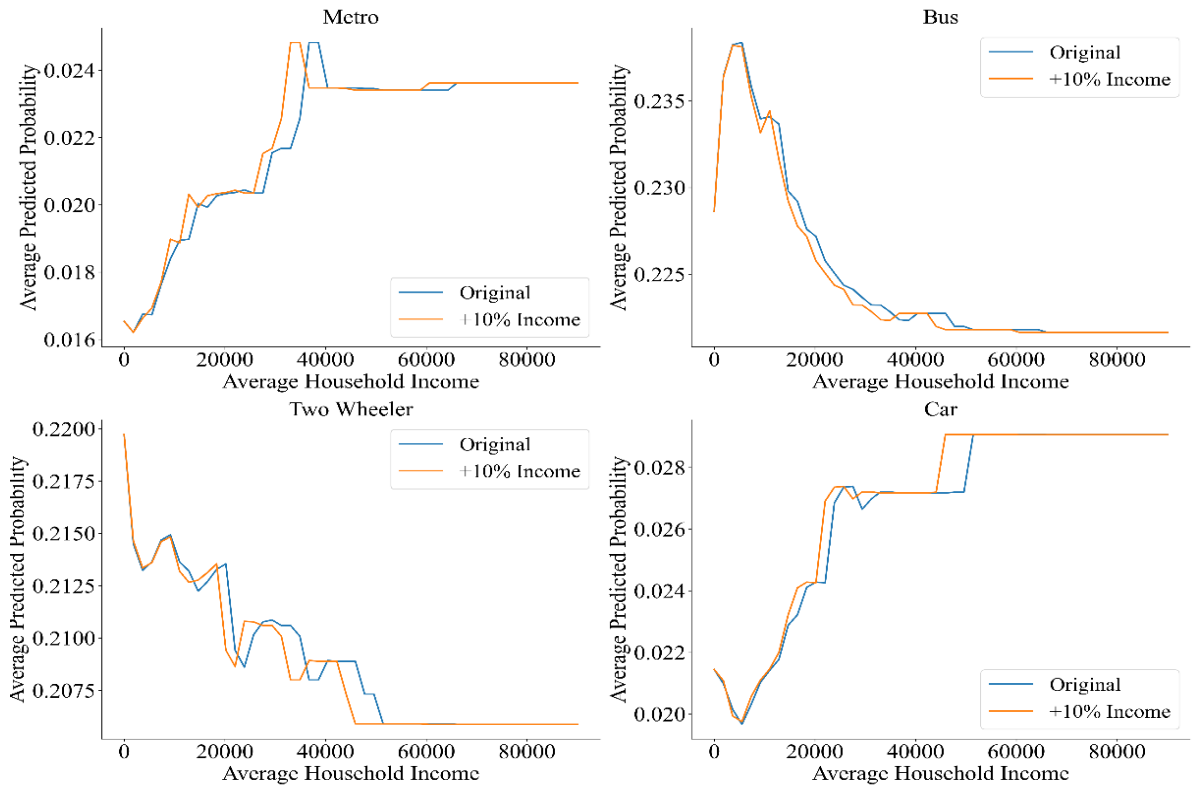

**Figure 23. ICE plots for Random Forest model between average household income and predicted probabilities for metro, bus, two-wheeler and car.**

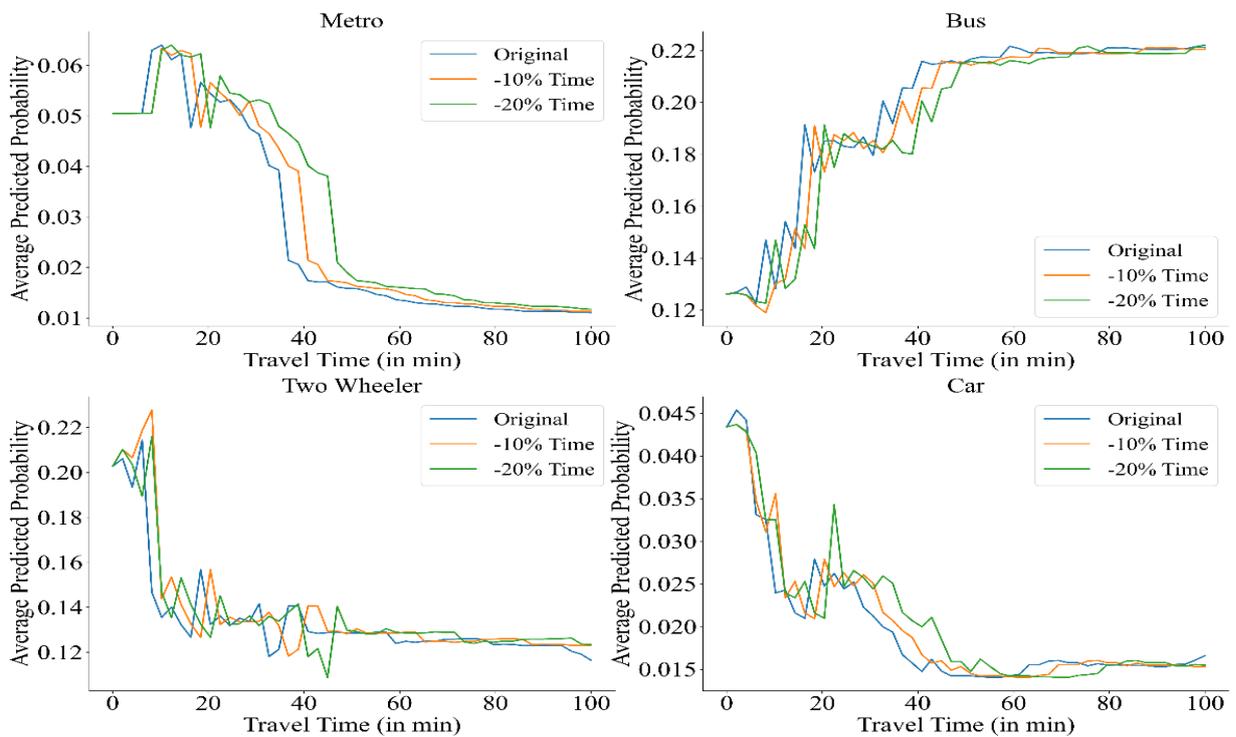

**Figure 24. ICE plots for Random Forest model between travel time and predicted probabilities for metro, bus, two-wheeler and car.**

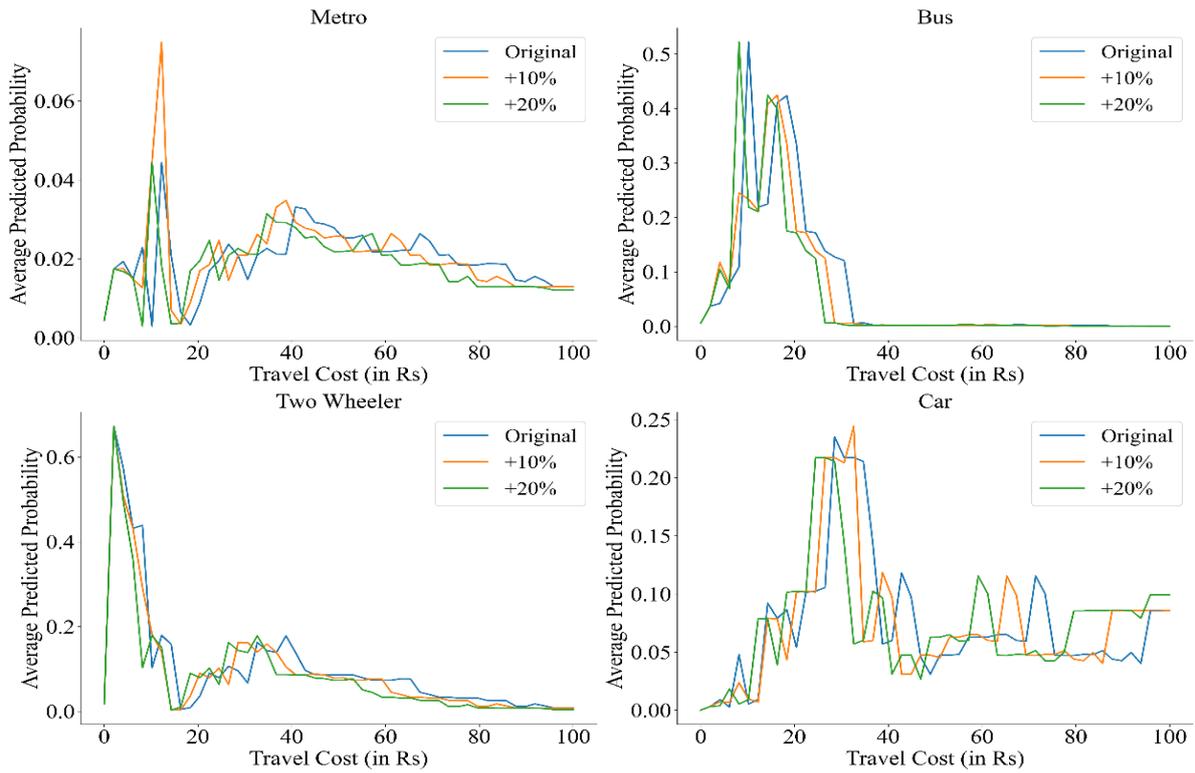

.**Figure 25. ICE plots for XGBoost model between travel cost and predicted probabilities for metro, bus, two-wheeler and car.**

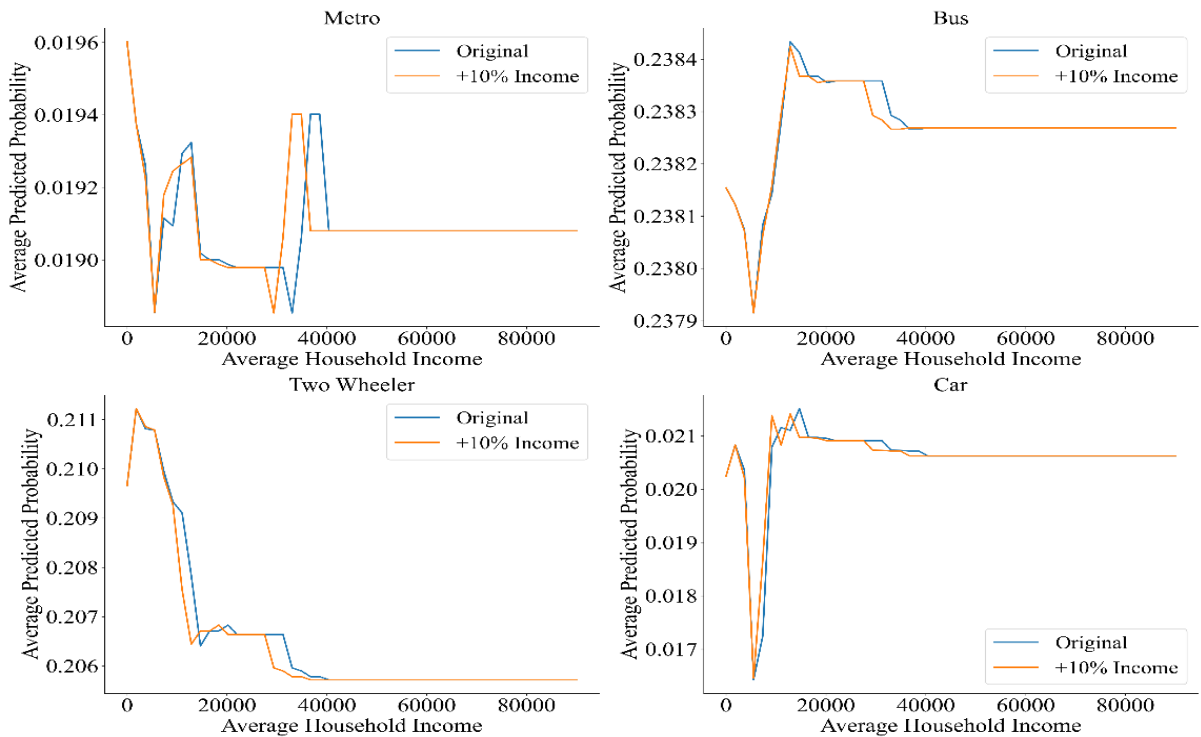

**Figure 26. ICE plots for XGBoost model between average household income and predicted probabilities for metro, bus, two-wheeler and car**

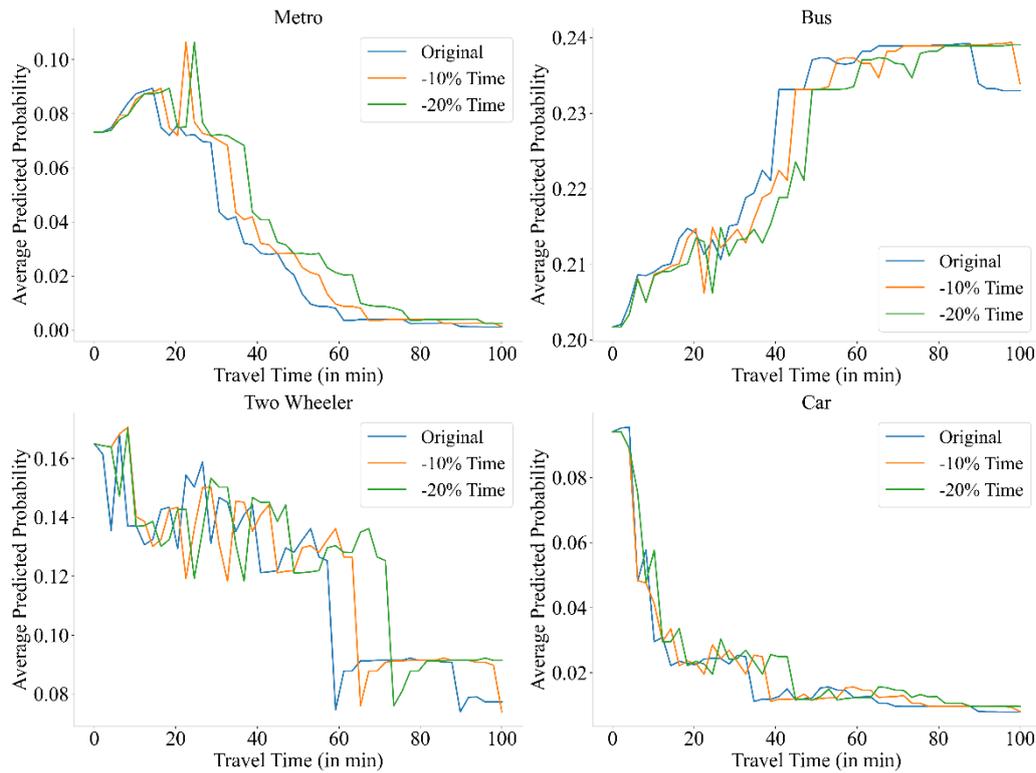

**Figure 27. ICE plots for XGBoost model between travel time and predicted probabilities for metro, bus, two-wheeler and car.**

In case of average ICE plots, a random forest model shows more variations in the predictions due to its structure which provides output by averaging across multiple decision trees whereas XGBoost show smoother change due to its gradient boosting approach. In the case of the metro, the predicted probability decreases with increase in travel cost showing a general sensitivity to price change among the individuals. As the travel cost increases beyond INR 100, the ICE lines flatten, possibly due to fewer data points at higher cost or reaching a saturation point in terms of price sensitivity which indicates that increase in cost prohibits individuals to use metro but after a certain threshold, the individuals who are using it are less sensitive to price change. For bus, a strong price sensitivity is observed among its users which is indicated by sharp decrease in the predicted probability lines when the travel cost increases. Similarly, there is a strong preference for two-wheelers when they are cheap mode options. However, unlike bus, a sharp decrease in predicted probability is not observed in the case of two wheelers suggesting less sensitivity to change in prices in general. However, in case of car, travel cost do not show a uniform effect on the probability of choosing car. The precited probability decreases with increase in cost but it increases after a threshold suggesting non-linear complex interactions with other features which is not captured by the ICE plot.

When average household income is considered, the predicted probability increases with increase in income till a certain point beyond which the curve flattens out suggesting that increase in income beyond a threshold, does not impact the probability of choosing metro as mode of commute. For scenario where 10% increase in average income of the household is constructed, a similar trend is observed which suggest the similar behaviour among the commuters even after an average increase of the household income. In case of bus, there is a sharp decrease in the probability of choosing bus following increase in average household income which suggest that individuals with lower household income have preference for the bus as mode of commute or they may be a captive user with very less option to switch to other mode. The trend is similar for the scenario where 10% increase in average household income is constructed suggesting that a small change in income do not substantially change the preference for bus which is intuitive as income to expenditure ratio is not going to improve substantially for this small change in income.

As travel time increase, the probability of choosing metro decreases indicating the preference of metro for shorter trips. As the travel time is hypothetically reduced under the constructed scenarios, the probability of choosing metro increases, showing an elasticity to travel time improvements. In case of bus, the predicted probability increases till a travel time of 60 minutes beyond which the curve stabilises. The decrease in travel time across the scenarios shows an increase in the probability of choosing bus. This result provides insights into the consistent behaviour of bus users possibly preferring low cost bus services over longer commute time or due to the constraints of routes and schedules which is maintained only by bus. There is decrease in the predicted probability of choosing two wheelers with increase in travel time, indicating that two wheelers are preferred for shorter trips and any improvement in travel time for shorter trips is more pronounced than longer trips. Similar to two wheelers, the probability of choosing car decreases as travel time increases. The reduction in travel time offered by the hypothetical scenarios have a little impact on the probability of choosing car suggesting the influence of travel time do not have much impact on the decision to use car by the individuals. A similar trend in the ICE plot between predicted probability and travel time is observed across all the modes for both the models.

Thus, ICE plots can be used to calculate marginal effects by examining the gradient of each line, which traces the impact of changing a feature on the predicted probability of selecting a specific transport option. For travel cost, individual ICE plots often show a pronounced negative marginal effect, with steeper slopes indicating a decrease in the probability of choosing cost-sensitive modes like the Bus as costs increase. For travel time, these individual plots typically illustrate a positive marginal effect: as time decreases, the probability of choosing time-sensitive modes like the Metro tends to increase.

When considering average ICE plots for travel cost, the aggregated effect across all observations suggests a consistent negative average marginal effect, as higher costs generally lead to a lower probability of choosing most transport modes. Similarly, average ICE plots for travel time display an average positive marginal effect: reduced travel times are associated with a higher average probability of selection across modes, reflecting a general preference for time-savings in transport decisions.

The ICE values for each instances can be averaged across all instances to calculate the average change in the probability of choosing a particular class which is essential to infer macro level trend with respect to the mode options. The average change in predicted probability is given by the following formula.

$$Average\ change\ = \frac{1}{N}\sum_{i=1}^{N} ICE_{y_i} - ICE_{x_i} \qquad 25$$

$x_i$ = Original feature values

$y_i$ = Modified feature values

$N$ = Number of instances in the dataset

Table 10 presents the average change in the predicted probability of different travel modes across each feature specific scenario.

**Table 10. Change in predicted probability for Random Forest and XGBoost**

| | Increase in travel cost | | Increase in Average Household income | Decrease in travel time | |
|---|---|---|---|---|---|
| | 10% | 20% | 10% | 10% | 20% |
| **Random Forest** | | | | | |

| Metro | -0.04% | -0.08% | 0.02% | 0.16% | 0.41% |
|---|---|---|---|---|---|
| Bus | -0.66% | -0.74% | -0.03% | -0.32% | -0.57% |
| Two Wheeler | -0.16% | -0.43% | 0.03% | -0.22% | -0.38% |
| Car | 0.93% | 1.70% | -0.01% | 0.12% | 0.39% |
| **XGBoost** | | | | | |
| Metro | 0.1% | -0.17% | 0.00% | 0.42% | 0.81% |
| Bus | -0.34% | -0.83% | 0.00% | -0.08% | -0.21% |
| Two Wheeler | -0.48% | -0.88% | 0.01% | -1.08% | -2.21% |
| Car | 2.34% | 4.11% | -0.01% | 0.25% | 0.52% |

As shown in table 10, for both the models, as the travel cost increases, the likelihood of choosing public transit modes like metro and bus decreases. This decrease is more pronounced for bus compared to metro which shows that bus users are more sensitive to change in travel cost. The probability of choosing private modes of transport (Two-wheeler and car) generally increases suggesting that the commuters using private modes of transport are less sensitive to increase in cost.

For change in average household income by 10%, the change in predicted probability across each model and each classes is relatively small suggesting that within the income range considered in this scenario, a moderate increase in income does not incentivise commuters to shift their mode of transportation. This also suggest that compared to income, other factors influence the model more.

With reduction in travel time, the probability of choosing metro and car across both the models increases and the preference for Bus and Two-Wheeler decreases. One of the plausible reason may be that the decrease in travel time across all modes deem the mode choices like

Buses and Two-Wheelers less competitive. Also, the result suggest that travel time does not play major role in the decision to choose bus or two-wheeler as the commute mode.

## 7. Discussion and Conclusions

Understanding the travel behaviour of marginalised population is crucial for equitable transportation planning (Barri et al., 2022). This understanding enables the planners to identify the factors which influences their mode choice behaviour. Accurately knowing such factors is a pre-requisite to introduce policy measures that enhances the equity of public transportation. This research have utilised MNL and ML models to understand the travel behaviour of such population and compared the predictability and behavioural aspect of each model.

The random forest model, among all the other models used on this dataset showed a superior predictive performance on both the training and testing data due to its ability to handle complex, non-linear relationship between variables due to its ensemble nature which combines multiple decision trees for the prediction purpose. Extreme gradient boosting and Decision trees perform on the training dataset exceptionally well (85% accuracy) but exhibits average performance on the testing dataset which shows the case of overfitting for both the models. SVM exhibits an above average performance on both the training (0.747) and testing dataset (0.543). While MNL model do not show predictive performance as high as Random Forest model but the accuracy on both the training (0.674) and testing data (0.616) for the MNL model is similar which shows the capability of these models to generalise on unseen data and therefore, the global nature of these models for classification task is better than the machine learning models. This generalisability is crucial for policy analysis, as it suggests that MNL models can reliably predict outcomes in various scenarios, making them valuable for long-term urban planning and policy development.

The behavioural interpretation of MNL model is its strength due to the statistical nature of such models. The results of the parameter estimates explains the behaviour of individuals in choosing a particular mode and how it varies for each feature used in the model. The economic underpinnings of the MNL model helps in understanding the economic interpretations using methods like Willingness to pay, Consumer surplus and direct as well as indirect elasticities. Two different methods have been employed to interpret the results the machine learning models- Feature importance and Individual conditional expectation plots. While feature importance summarises the contribution of each feature of interest in the model's predictive output, ICE plots explains in a granular fashion the mode choice behaviour of each individual to the change in the value of a particular feature variable. As shown in the feature importance plots, the travel cost emerges as the predominant feature followed by features like travel time and vehicle availability across all the machine learning models. Using this insight, ICE plots for two tree based models (Random forest and XGB) is used to understand how the individuals change the mode choice decision making when there is a change in the travel cost, travel time and the household income. While individual ICE plots are used to show the variability in mode choice behaviour of each individuals, the average ICE plots which is constructed by average the individual ICE for the entire dataset show a broader trend and patterns in the dataset. While similar results can be derived using choice simulation on each individual datapoint using the probabilistic framework of MNL model, the ICE plots are more granular in nature as the feature variable changes over a significant range, thus capturing subtle nuances of how changing a particular feature can impact the mode choice decision making of individuals. Recent studies have exploited interpretability techniques to explain travel behaviour (Zhao et al., 2018, Kashifi et al., 2022, Mart´ın-Baos et al., 2023).

The results of this study have important implications for urban planning, policy development, and the creation of travel demand models. As policymakers strive to address inequalities in access to transportation and improve community participation, this research offers insights into how people choose their mode of travel. The study introduces a framework that shows how machine learning methods can make predictions about travel mode choices at par with statistical models. However, a key question is which model should be used in practice. There's a trade-off between accuracy and understandability. Based on our findings and a thorough review of existing research, we suggest a systematic approach. First, we should train different machine learning and statistical models. Then, we should carefully evaluate how well each model performs using various measures. If the differences in predictive performance are small, it's better to choose a model that's easy to understand. Next, we should use tools to dig deeper into how the chosen model works and consult with experts to improve our understanding of it. Ultimately, we should trust the model's results if they align with what we already know from research and real-world observations.

A hybrid modelling approach is an important area of interest for various researchers in this field. Such approach would combine the superior predictive ability of machine learning models along with the behaviour interpretability of MNL models. Current packages to estimate MNL models do not have automated feature selection approach. Therefore, a sequential hybrid modelling approach which first selects feature using the feature selection method of Random forest and then apply MNL models to these selected feature would combine the strength of both the models for a robust modeling approach (Prinzie and Poel, 2008).

There are few limitations of this study which require further work. One area of future research is the comparison with similar field survey based travel dataset from developing countries to explore the predictability and generalisability of these models. Additionally, other machine learning models, specifically, neural network based models can be used in similar dataset to explore their generalisability and interpretability on such dataset. Additionally, since certain parameters of the MNL model such as travel cost and travel time are non-monotonic, behavioural interpretation becomes more robust. Integrating monotonicity constraints to tree based models is an important area of research for achieving more realistic behavioural interpretation.

**Table 11. Comparative framework of MNL and ML based on the findings.**

| Aspects | MNL | ML |
|---|---|---|
| Model formulation | Based on utility maximisation and probabilistic choice framework | Tree based (DT), ensemble methods (RF, XGB) |
| Model type | Parametric, statistical | Non-parametric , algorithmic |
| Input data | Structured input data. Categorical variables often transformed into dummy/indicator variables | Tolerant of unstructured data |
| Prediction type | Probabilistic in nature | Label type output which can be converted to probabilities |
| Optimisation method | Maximum likelihood based | Bootstrapping and boosting |
| Hyperparameter optimisation | Typically involves selecting best fit models using the metrics like likelihood ratio test, AIC (Akaike information criteria)/BIC (Bayesian information criteria | Involves tuning a range of parameters like number of trees, depth of tree etc using grid search, random search or Bayesian methods |
| Individual level prediction | Choice probabilities at individual level | Rule based predictions by decision trees. |

| | | Ensemble methods aggregate the predictions of multiple decision trees |
|---|---|---|
| Aggregate level prediction | Aggregate probabilities through simulation | Direct aggregation possible; ensemble models can improve accuracy |
| Variable importance | Coefficient size and sign after model estimation. | Built-in feature importance metrics, especially in RF and XGB. |
| Variable effect | Direct and interpretable effects via parameter estimates. | Methods like partial dependence plots and Individual conditional expectation plots |
| Marginal effects | Directly calculated from the model parameters and the probability formulas. | Indirect; requires techniques such as ICE plots or derivatives of prediction functions. |

Table 11 provides a concise comparative analysis, illustrating the fundamental differences between MNL and ML models in terms of their conceptual framework, data handling, prediction methods, and analytical capabilities, thereby delineating their respective advantages and applications in travel behavior analysis. Therefore, an effective modeling of travel mode choice is hinged upon the aspects discussed in table 11 and selection of the appropriate model should follow categorically the model's performance in these aspects.

**Declaration of competing interests**

The authors declare that they have no known competing financial interests or personal relationships that could have appeared to influence the work reported in this paper.

**Acknowledgements**

The authors are grateful to the Ministry of Earth Sciences (MoES), Government of India, for supporting the PhD dissertation research of the first author. The authors are grateful to Prof. Raman Srikanth, principal investigator of the project, for his exceptional support and invaluable guidance. Additionally, the authors also wish to extend their heartfelt thanks to Dr. Tejal Kanitkar (NIAS, Bengaluru) and Dr. Harikrishnan NB (BITS Pilani, Goa campus) for their significant inputs and suggestions.

**Funding**

This research work is part of a larger research project being implemented by NIAS Bengaluru under a Ministry of Earth Sciences (MOES, GOI) Grant (MoES/ 16/15/2011-RDEAS (NIAS) dated May 22, 2018) on the theme "*to understand the Interaction between components of Earth and Human Systems at various Spatial and Temporal Scales*".